\documentclass[lettersize,journal]{IEEEtran}
\usepackage{amsmath,amsfonts}
\usepackage{algorithmic}
\usepackage{algorithm}
\usepackage{array}
\usepackage[caption=false,font=normalsize,labelfont=sf,textfont=sf]{subfig}
\usepackage{textcomp}
\usepackage{dblfloatfix}
\usepackage{url}
\usepackage[hidelinks]{hyperref}
\newcommand{\orcid}[1]{\textsuperscript{\href{https://orcid.org/#1}{\includegraphics[width=8pt]{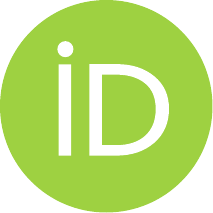}}}}
\usepackage{verbatim}
\usepackage{graphicx}
\usepackage{tabularx}
\newcolumntype{C}{>{\centering\arraybackslash}X} % centered text in C columns
\usepackage{cite}
\usepackage{longtable}
\usepackage{colortbl}
\usepackage{xcolor}
\usepackage{booktabs}
\usepackage{hhline}
\usepackage{multirow}
\graphicspath{{figures/}{experiments/}}
\usepackage{makecell}

\makeatletter
\newcommand{\thickhline}{%
    \noalign {\ifnum 0=`}\fi \hrule height 0.7pt
    \futurelet \reserved@a \@xhline
}

\newcommand{\tabincell}[2]{
    \begin{tabular}{@{}#1@{}}
        #2
    \end{tabular}
}

\newcolumntype{"}{@{\vrule width 1pt \hskip\tabcolsep}}
\makeatother

\begin{document}

\title{Deeper into Self-Supervised Monocular\\ Indoor Depth Estimation}

\author{Chao Fan\orcid{0000-0001-9589-7567}, Zhenyu Yin\orcid{0000-0001-5403-3922},~\IEEEmembership{Member,~IEEE}, Yue Li\orcid{0000-0002-0222-9859}, and Feiqing Zhang\orcid{0000-0003-4636-9607}
        % <-this % stops a space

\thanks{Chao Fan, Yue Li and Feiqing Zhang are with the University of Chinese Academy of Sciences, Beijing 100049, China, also with the Shenyang Institute of Computing Technology, Chinese Academy of Sciences, Shenyang 110168, China, and also with the Liaoning Key Laboratory of Domestic Industrial Control Platform Technology on Basic Hardware \& Software, Shenyang 110168, China (e-mail: fanchao18@mails.ucas.ac.cn; liyue161@mails.ucas.edu.cn; zhangfeiqing17@mails.ucas.ac.cn).}% <-this % stops a space
\thanks{Zhenyu Yin is with the Shenyang Institute of Computing Technology, Chinese Academy of Sciences, Shenyang 110168, China, and also with the Liaoning Key Laboratory of Domestic Industrial Control Platform Technology on Basic Hardware \& Software, Shenyang 110168, China (e-mail: congmy@163.com).}}

\markboth{Journal of \LaTeX\ Class Files,~Vol.~14, No.~8, August~2021}%
{Shell \MakeLowercase{\textit{et al.}}: A Sample Article Using IEEEtran.cls for IEEE Journals}

\IEEEpubid{0000--0000/00\$00.00~\copyright~2021 IEEE}

\maketitle

\begin{abstract}
Monocular depth estimation using Convolutional Neural
Networks (CNNs) has shown impressive performance in outdoor driving scenes. However, self-supervised learning of indoor depth from monocular sequences is quite challenging for researchers because of the following two main reasons. One is the large areas of low-texture regions and the other is the complex ego-motion on indoor training datasets. In this work, our proposed method, named IndoorDepth, consists of two innovations. In particular, we first propose a novel photometric loss with improved structural similarity (\text{SSIM}) function to tackle the challenge from low-texture regions. Moreover, in order to further mitigate the issue of inaccurate ego-motion prediction, multiple photometric losses at different stages are used to train a deeper pose network with two residual pose blocks. Subsequent ablation study can validate the effectiveness of each new idea. Experiments on the NYUv2 benchmark demonstrate that our IndoorDepth outperforms the previous state-of-the-art methods by a large margin. In addition, we also validate the generalization ability of our method on ScanNet dataset. Code is availabe at \emph{\href{https://github.com/fcntes/IndoorDepth}{https://github.com/fcntes/IndoorDepth}}.
\end{abstract}


\section{Introduction}\label{sec:intro}
\IEEEPARstart{D}{epth} estimation plays a significant role in the development of 3D
computer vision, especially for the applications in autonomous driving
\cite{Autonomous1,Autonomous2,Autonomous3,KITTI0}, 3D reconstruction \cite{3D1,3D2}, virtual reality \cite{VR1,VR2,VR3}, etc. In recent
years, there has been growing interest in monocular depth estimation
based on deep learning. Compared with supervised learning approaches
\cite{Supervised1,Supervised2,Supervised3,Supervised4,DORN} which need a large amount of ground-truth depth, self-supervised
learning methods \cite{MD1,MD2,Semantic1} use the photometric consistency as the
supervisory signal. Therefore, it is more attractive for researchers to
adopt self-supervised methods to avoid the use of expensive sensors
(e.g. LiDAR). Some recent self-supervised methods \cite{DDV,PackNet,Fine-grained} have achieved
significant performance in outdoor scenes, such as the KITTI dataset
\cite{KITTI} and Make3D dataset \cite{Make3D}.

% \newgeometry{bottom=5cm}   
Although such self-supervised methods have been able to obtain
high-quality depth maps in outdoor scenes, they usually collapse easily
in indoor scenes like NYUv2 dataset \cite{NYUv2} and ScanNet dataset \cite{ScanNet}.
It is still challenging to obtain accurate indoor depth estimation
results for self-supervised methods due to the following reasons. (1) Lack of textures: large low-texture regions make the network difficult
to train because of the ambiguity of photometric loss in these regions.
Specifically, the image gradients are close to zero which further causes
zero gradients for both depth and pose estimation \cite{feature-metric}.
(2) Challenging pose estimation: the main reason why it becomes
difficult to predict accurate camera poses is that the ego-motion is
more complicated since indoor images are usually captured by hand-held
devices. Compared with camera poses in outdoor environments where
translations are dominant, indoor poses contain frequent rotations which
have been proved to behave as noise to self-supervised depth learning
\cite{ARN}.\IEEEpubidadjcol

In this paper, we propose \textbf{IndoorDepth}, a self-supervised
monocular indoor depth estimation method by combining two innovations.
First, in order to mitigate the issue of low-texture regions, we propose
a novel photometric loss, which achieves a more suitable discrimination
of structural similarity by an improved structural similarity (\text{SSIM}) \cite{SSIM} function.
Specifically, we make it more sensitive to subtle differences in
structure by reducing the structural similarity index. This allows us to
obtain more reasonable structural similarity index for low-texture
regions in contrast to the original very high and undiscriminating index.

Second, we propose a deeper pose network trained by multiple photometric
losses which can further alleviate the problem of inaccurate rotation prediction. Following the
iterative strategy in MonoIndoor \cite{MonoIndoor}, we successfully construct a
deeper pose network with more residual pose blocks. But we do not simply
stack more residual blocks, because it has been proved that the
performance does not significantly improve or even becomes worse when
the number of residual blocks is more than one \cite{MonoIndoor}. In order to make
all the added residual pose blocks work, we use multiple photometric
losses at different stages to train the pose network. Subsequent
experiments can demonstrate that such a strategy leads to better
results.

Our main contributions can be summarized as follows:

\begin{itemize}
\item A novel photometric loss with an improved \text{SSIM} function which helps us to obtain more reasonable structural similarity index for low-texture regions.
\item  A deeper pose network trained by multiple photometric losses at
different stages which can further mitigate the issue of inaccurate rotation prediction and in turn lead to better depth prediction.
% \newpage
\item  Extensive experiments on NYUv2 \cite{NYUv2} dataset show that our results outperform the previous state-of-the-art method by a large margin.
\end{itemize}

\section{Related Work}

In this section, we review monocular depth estimation methods including two categories: supervised learning and self-supervised learning.
\subsection{Supervised Monocular Depth Estimation}

With the development of Convolutional Neural Networks (CNNs), lots of
supervised methods for monocular depth estimation have been
proposed. Eigen et al. \cite{Eigen1} first implemented monocular depth
estimation using a multi-scale approach based on deep learning.
Afterward, many works have focused on using better network structures
\cite{Supervised3,Supervised6,DFDepth,Pyramid-Depth,CORNet} or loss functions \cite{Supervised9,Supervised10,DORN} to improve the
performance of depth estimation. Such supervised methods usually can
achieve better performance than self-supervised methods for the same
dataset. However, the requirement of ground-truth depth limits the
application scenarios of these supervised learning approaches.
% \newpage

\subsection{Self-Supervised Monocular Depth Estimation}

Since self-supervised monocular depth estimation only requires stereo
image pairs or monocular videos as input, lots of researchers have tried
to improve its performance on different datasets. Here we only focus on
self-supervised models trained on monocular sequences. Zhou et al.
\cite{ZhouT} first achieved self-supervised monocular depth estimation by
simultaneously training a depth network and a pose network. Godard et
al. \cite{MD2} adopted a minimum reprojection loss, a full-resolution
multi-scale sampling method and an auto-masking loss to obtain better
predictions. Self-supervised learning approaches require a strict assumption that all objects except the camera remain stationary. To handle moving objects such as cars and pedestrians in outdoor scenes, Li et al. \cite{motionnet} proposed a motion network which can predict both ego-motion and objects motion. Recently, by the novel design of network architectures \cite{PackNet,HRDepth}, better loss functions \cite{FD-Photometric}, multi-frame during test
\cite{DontForget,ManyDepth} and cross-domain knowledge learning
\cite{Fine-grained}, self-supervised learning methods have achieved impressive
performance on many outdoor datasets.

However, most of these methods failed to obtain satisfactory results in
the challenging indoor scenes due to more complicated ego-motion and
large textureless regions. Zhou et al. \cite{MovingIndoor} employed an
optical-flow based training paradigm which reduced the difficulty of
self-supervised learning. They used the flow field generated by an
optical flow estimation network as supervision. Yu et al. \cite{P2Net}
innovatively adopted patch-based matching instead of pixel-based
matching to obtain a more reasonable photometric loss. In order to further constrain low-texture areas, they proposed a planar consistency loss according to the assumption that homogeneous regions tends to be planar regions. Li et al.
\cite{StructDepth} leveraged the structural regularities including
Manhattan normal constraint and co-planar constraint to achieve better
depth maps. PLNet \cite{PLNet} further used the line and plane clues in monocular sequences to
enhance the depth prediction. Recently, MonoIndoor++ \cite{MonoIndoor++} achieved
state-of-the-art performance by using a depth factorization module and a
residual pose estimation module. Our proposed model based on PLNet,
named IndoorDepth, is extended with our two innovations.

\section{Method}

In this section, we first introduce our basic model, PLNet, and then
give detailed descriptions of our method, IndoorDepth, including a novel
photometric loss with improved \text{SSIM} and a deeper pose network with
multiple photometric losses. Finally, we obtain the overall loss
function by combining the basic model and our loss innovations.
\subsection{Basic Model\label{sec:3A}}

Following Monodepth2 \cite{MD2}, lots of self-supervised monocular indoor
depth estimation approaches have been proposed. PLNet \cite{PLNet} is the
method that makes the best use of plane and line priors and results in a
very well-developed loss function. Here we briefly review PLNet containing a DepthNet and a PoseNet, as it
is our backbone model for depth prediction.

Generally, a training sample consists of a target frame \(I_{t}\) and an
adjacent frame \(I_{t^{\prime}}\). Unlike most self-supervised methods that
directly estimate the normalized disparity map of \(I_{t}\), PLNet first
makes the DepthNet output the local plane coefficients and then converts
them to depth \(D_{t}\) by the following equation:
\begin{equation}
    1/D_{t} = {co}^{T}K^{-1}p_{t}
    \label{eq:1}
\end{equation}
where the vector \(co = ({co}_{x},\ {co}_{y},\ {co}_{z})\) denotes the planar
coefficient, \(K\) is the camera intrinsics and \(p_{t}\) is the pixel coordinate in \(I_{t}\). During training, the PoseNet
outputs the relative pose \(T_{t \rightarrow t^{\prime}}\) between \(I_{t}\)
and \(I_{t^{\prime}}\). Then we can formulate the transformation equation to
warp one view into another,
\begin{equation}
    p_{t^{\prime}}\sim KT_{t \rightarrow t^{\prime}}D_{t}{(p_{t})K}^{- 1}p_{t}
    \label{eq:2}
\end{equation}
where \(p_{t^{\prime}}\) denotes the pixel in the source view \(I_{t^{\prime}}\).
The re-projected image is denoted as \(I_{t^{\prime} \rightarrow t}\), which
should perfectly match the target image \(I_{t}\).

Next, we need to construct the loss function to train both the DepthNet and
the PoseNet. Similar to \cite{MD1,MD2}, the photometric reprojection error
between the target view \(I_{t}\) and synthesized view
\(I_{t^{\prime} \rightarrow t}\) is a weighted combination of \text{SSIM} and L1,
which is defined as:
\begin{equation}
    L_{ph} = \frac{\alpha}{2}( 1 - \text{SSIM}( I_{t},\ I_{t^{\prime} \rightarrow t} ) ) + (1 - \alpha)\| I_{t} - I_{t^{\prime} \rightarrow t}\|_{1} \label{eq:3}
\end{equation}
where \(\alpha = 0.85\). We also follow PLNet to compute the smoothness
term on the planar coefficients since it makes more sense to regularize the smoothness on the planar coefficients than on the disparity.
The smoothness regularization term is defined as:
\begin{equation}
    L_{sm} = \sum_{i \in \{ x, y, z\}}^{}{(\left| \partial_{u}{co}_{i}^{*} \right|e^{- \left| \partial_{u}I_{t} \right|} + \left| \partial_{v}{co}_{i}^{*} \right|e^{- \left| \partial_{v}I_{t} \right|})}
\end{equation}
where \({co}_{i}^{*} = {co}_{i}/\overline{\left| {co}_{i} \right|}\) is the
absolute-mean-normalized coefficient to discourage its shrinking. Note
that in order to make sure the depth in Eq. (\ref{eq:1}) is positive, we use the
same processing for \(co\) as that in PLNet. Furthermore, we use the
plane consistency loss \(L_{pc}\) and the linear consistency loss
\(L_{lc}\) to enforce plane regularization and linear regularization,
respectively. They are defined as:
\begin{equation}
    L_{pc} = \frac{1}{N_{p}}\sum_{i = 1}^{N_{p}}\left| \overrightarrow{A_{i}B_{i}} \times \overrightarrow{A_{i}C_{i}} \cdot \overrightarrow{A_{i}D_{i}} \right|
\end{equation}

\begin{equation}
    L_{lc} = \frac{1}{N_{l}}\sum_{i = 1}^{N_{l}}\left| \overrightarrow{E_{i}F_{i}} \times \overrightarrow{E_{i}G_{i}} \right|
\end{equation}
where \(N_{p}\) denotes the number of 4-point sets selected from pseudo-planes and \(N_{l}\) denotes the number of 3-point sets selected from line segments. A, B, C and D are the corresponding
3D points randomly sampled from a pseudo-plane. E, F and G are the 3D
points randomly sampled from a line segment. The overall loss \(L\) of
our baseline is a combination of the above loss functions, i.e.
\begin{equation}
    L = L_{ph} + {\alpha_{sm}L}_{sm} + {\alpha_{pc}L}_{pc} + {\alpha_{lc}L}_{lc} \label{eq:7}
\end{equation}
where \(\alpha_{sm}\), \(\alpha_{pc}\) and \(\alpha_{lc}\) are
hyper-parameters to balance different regularization terms.

\begin{figure*}[h!tb]
    \centering
    \includegraphics[width=.9\textwidth]{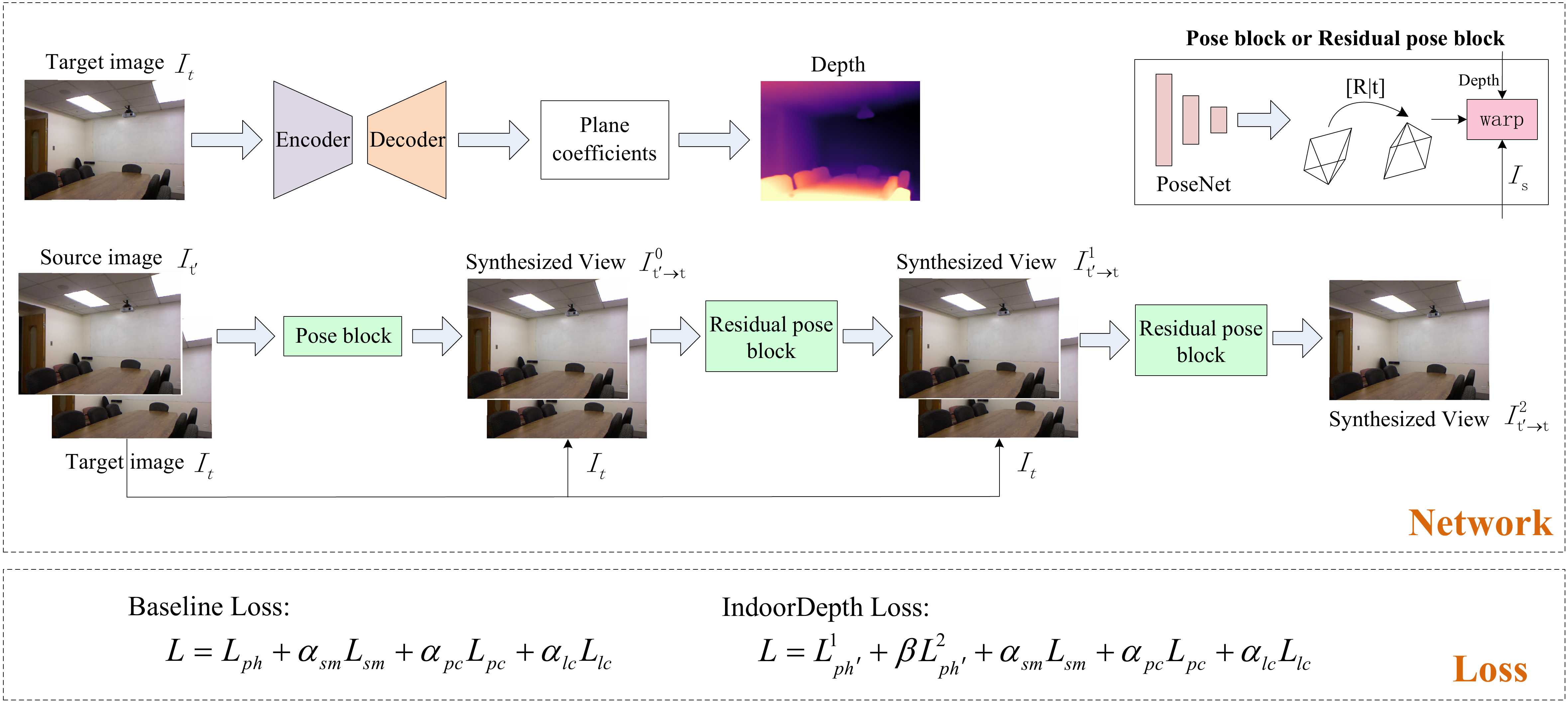}
    \caption{The overall framework of our IndoorDepth. The input of our depth network is the target image \(I_{t}\) and the output is the corresponding depth map. Training images \(I_{t}\) and \(I_{t^{\prime}}\) are passed into the
    pose network, including three blocks, to finally predict the relative
    pose between \(I_{t}\) and \(I_{t^{\prime} \rightarrow t}^{1}\) and obtain
    the synthesized view \(I_{t^{\prime} \rightarrow t}^{2}\). \(I_{s}\) denotes the corresponding source images at different stages, including \(I_{t^{\prime}}\), \(I_{t^{\prime} \rightarrow t}^{0}\) and \(I_{t^{\prime} \rightarrow t}^{1}\). In the loss function, we replace the photometric loss \(L_{ph}\) in the baseline with our novel photometric loss and employ multiple photometric losses to constrain the network.\label{fig:overview}}
\end{figure*}

As mentioned in Section \ref{sec:intro}, lots of self-supervised learning methods fail
to obtain satisfactory depth predictions in indoor environments. The
main reasons for this are the large number of low-texture areas in
indoor scenes and the more complex ego-motion. To address these issues,
we propose IndoorDepth, as shown in Fig. \ref{fig:overview}, to predict more accurate
depth maps by using our novel photometric loss with improved \text{SSIM} and
deeper pose network with multiple photometric losses. More details of
our contributions can be found in the following sections.

\subsection{Photometric Loss with Improved \text{SSIM}\label{sec:3B}}
From Eq. (\ref{eq:7}), we can find that the photometric loss \(L_{ph}\) is the
main loss function, and the others are just different regularization
terms. Therefore, in order to further improve the performance of
self-supervised depth estimation by using novel training loss, we try to
focus on the photometric loss \(L_{ph}\). As defined in Eq. (\ref{eq:3}), the
photometric loss \(L_{ph}\) is a combination of \text{SSIM} loss and L1 loss.
On one hand, the L1 loss is used to preserve colors and luminance
\cite{SSIML1}; on the other hand, the \text{SSIM} loss attempts to preserve
the structural information.

However, there are a large number of low-texture areas in indoor
environments. The mean value of L1 losses in a low-texture region is
always too small, which means that incorrect matching cannot be
appropriately penalized. To handle the issue of
low-texture regions, we consider relying more on the structural
similarity provided by \text{SSIM} to penalize incorrect estimation of depth
and pose. In fact, another reason why we explore the structural
information is that \text{SSIM} has relatively large room for improvement as a
metric of structural similarity, while the form of LI loss is already
fixed. Here we separate out the \text{SSIM} part \(L_{\text{ssim}}\) from Eq. (\ref{eq:3}) to obtain the following equation:
\begin{equation}
    L_{\text{ssim}} = \frac{\alpha}{2}( 1 - \text{SSIM}(I_{t},\ I_{t^{\prime} \rightarrow t}))
    \label{eq:8}
\end{equation}
where \(I_{t}\) is the target view and \(I_{t^{\prime} \rightarrow t}\) is
the synthesized view. \(\text{SSIM}\) is a structural similarity index to
evaluate the structure similarity between \(I_{t}\) and \(I_{t^{\prime} \rightarrow t}\). Given
two images \(I_{x}\) and \(I_{y}\), the original \(\text{SSIM}\) \cite{SSIM} is computed as:
\begin{align}
    \text{\text{SSIM}}( I_{x},\ I_{y} ) &= l( I_{x},\ I_{y} )cs( I_{x},\ I_{y} ) \notag \\
    l( I_{x},\ I_{y} ) &= \frac{2\mu_{x}\mu_{y} + C_{1}}{\mu_{x}^{2} + \mu_{y}^{2} + C_{1}} \notag \\
    cs( I_{x},\ I_{y} ) &= \frac{2\sigma_{xy} + C_{2}}{\sigma_{x}^{2} + \sigma_{y}^{2} + C_{2}} \label{eq:9}
\end{align}
where \(\mu_{x}\) denotes the mean value of \(I_{x}\) and \(\mu_{y}\)
denotes the mean value of \(I_{y}\). \(\sigma_{x}\) is the standard
deviation of \(I_{x}\) and \(\sigma_{y}\) is the standard deviation of
\(I_{y}\). \(\sigma_{xy}\) denotes the covariance of \(I_{x}\) and
\(I_{y}\). The \(\text{SSIM}\) is based on the luminance comparison function
\(l( I_{x},\ I_{y} )\) and the contrast and structure
comparison function \(cs( I_{x},\ I_{y} )\). The constant
\(C_{1}\) and \(C_{2}\) are added to avoid instability when
\(\mu_{x}^{2} + \mu_{y}^{2}\)
or \(\sigma_{x}^{2} + \sigma_{y}^{2}\) are close to zero. Specifically,
they are defined as:
\begin{align}
    C_{1} &= {(M_{1}L)}^{2} \notag \\
    C_{2} &= {(M_{2}L)}^{2}
\end{align}
where \(M_{1} = 0.01\) and \(M_{2} = 0.03\). \(L\) denotes the dynamic
range of the pixel values (255 for 8-bit grayscale images or 1 for
normalized images).

In this paper, we implement a novel photometric loss via improving
the original \(\text{SSIM}\). Our improved structural similarity function \({\text{SSIM}}^{\prime}\) is able to be more
sensitive to subtle differences by reducing the structural similarity
index. Given two original images \(I_{x}\) and \(I_{y}\), we first
compute the linearly transformed images \(I_{kx}\) and \(I_{ky}\) by the
following equation:
\begin{align}
    I_{kx} &= kI_{x} \notag \\
    I_{ky} &= kI_{y}
    \label{eq:11}
\end{align}
where \(k\) \((k > 1\)) is a hyper-parameter. Then the \({\text{SSIM}}^{\prime}\) is
defined as:
\begin{align}
    {\text{SSIM}}^{\prime}( I_{x},\ I_{y} ) &= l^{\prime}( I_{x},\ I_{y} ){cs}^{\prime}( I_{x},\ I_{y} ) \notag \\   
    l^{\prime}( I_{x},\ I_{y} ) &= \frac{2\mu_{kx}\mu_{ky} + C_{1}}{\mu_{kx}^{2} + \mu_{ky}^{2} + C_{1}} \notag \\
    {cs}^{\prime}( I_{x},\ I_{y} ) &= \frac{2\sigma_{kxky} + C_{2}}{\sigma_{kx}^{2} + \sigma_{ky}^{2} + C_{2}}
\end{align}
where \(\mu_{kx}\) is the mean value of \(I_{kx}\) and \(\mu_{ky}\)
denotes the mean value of \(I_{ky}\). \(\sigma_{kx}\) denotes the
standard deviation of \(I_{kx}\) and \(\sigma_{ky}\) denotes the
standard deviation of \(I_{ky}\). \(\sigma_{kxky}\) is the covariance of
\(I_{kx}\) and \(I_{ky}\). Table \ref{tab:SSIM} shows the differences of the
corresponding \(C_{1}\) and \(C_{2}\) in various structural similarities. Note that
\(C_{1}\) and \(C_{2}\) in \({\text{SSIM}}^{\prime}( I_{x},\ I_{y} )\)
are determined by the range of the pixel values in the input \(I_{x}\)
and \(I_{y}\), not \(I_{kx}\) and \(I_{ky}\). Meanwhile it is easy to
conclude that our \({\text{SSIM}}^{\prime}\) satisfies the following three
conditions similar to the original \(\text{SSIM}\).
\begin{enumerate}
    \item Symmetry:\(\ {\text{SSIM}}^{\prime}( I_{x},\ I_{y} ) = {\text{SSIM}}^{\prime}( I_{y},\ I_{x} )\)
    \item Boundedness: \({\text{SSIM}}^{\prime}( I_{x},\ I_{y} ) \leq 1\)
    \item Unique maximum: \({\text{SSIM}}^{\prime}( I_{x},\ I_{y} ) = 1\) if and only if \(I_{x} = I_{y}\).
\end{enumerate}
\begin{figure*}[!htb]
    \centering
    \includegraphics[width=.85\textwidth]{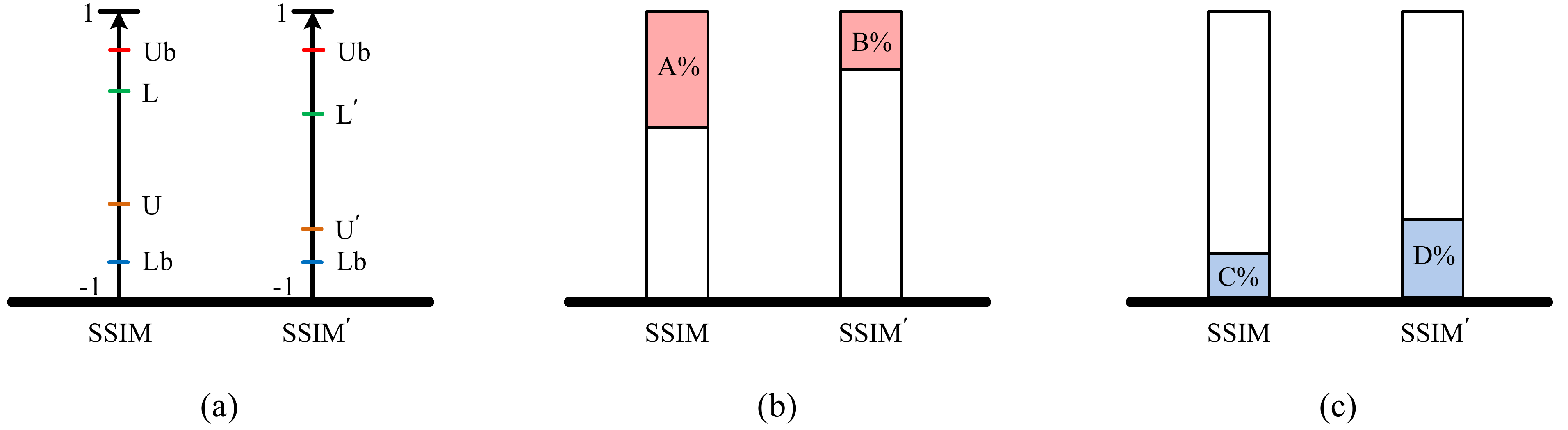}
    \caption{Analysis of the reasons why \({\text{SSIM}}^{\prime}\) is superior
    to \(\text{SSIM}\). (a) Multiple upper and lower bounds of structural
    similarity indexes. \(Ub\) and \(Lb\) denote the upper and lower bounds
    of the valid structural similarity indexes, respectively. For a sample
    of the same low-texture region, \(L\) and \(L^{\prime}\) are the lower bounds
    of \(\text{SSIM}\) and \({\text{SSIM}}^{\prime}\), respectively. For a sample of the same
    noisy region, \(U\) and \(U^{\prime}\) are the upper bounds of \(\text{SSIM}\) and
    \({\text{SSIM}}^{\prime}\), respectively. (b) Percentage of invalid indexes in a
    sample of the same low-texture region. The invalid indexes here denote
    the structural similarity indexes within the interval {[}\(Ub\), 1{]}.
    (c) Percentage of invalid indexes in a sample of the same noisy region.
    The invalid indexes here denote the structural similarity indexes within
    the interval {[}-1, \(Lb\){]}.\label{fig:analysis}}
\end{figure*}
\begin{table}[!tb]
    \caption{\(C_{1}\) and \(C_{2}\) in Different Structural Similarities.\label{tab:SSIM}}
    \centering
    % \small
    \renewcommand\arraystretch{1.7}
    \begin{tabularx}{.8\linewidth}{cCC}
        % \hline
        \toprule
        Structural similarity & $C_1$ & $C_2$ \\ 
        \midrule
        \(\text{SSIM}( I_{x},\ I_{y} )\) & \({(M_{1}L)}^{2}\) &
        \({(M_{2}L)}^{2}\) \\ %\hline
        \(\text{SSIM}( I_{kx},\ I_{ky} )\) & \({(M_{1} \cdot kL)}^{2}\) &
        \({(M_{2} \cdot kL)}^{2}\) \\ %\hline
        \({\text{SSIM}}^{\prime}( I_{x},\ I_{y} )\) & \({(M_{1}L)}^{2}\) &
        \({(M_{2}L)}^{2}\) \\
% \hline
        \bottomrule
    \end{tabularx}
\end{table}

In our supplementary material, we further prove the following three
conclusions:
\begin{equation}
    \text{SSIM}( I_{x},\ I_{y} ) = \text{SSIM}( I_{kx},\ I_{ky} )
    \label{eq:13}
\end{equation}
\begin{equation}
    {\text{SSIM}}^{\prime}( I_{x},\ I_{y} ) \leq \text{SSIM}( I_{x},\ I_{y} )
    \label{eq:14}
\end{equation}
\begin{equation}
    {\text{SSIM}}_{k_{1}}^{\prime}( I_{x},\ I_{y} ) \leq {\text{SSIM}}_{k_{2}}^{\prime}( I_{x},\ I_{y} )
    (k_{1} > k_{2}>1)
    \label{eq:15}
\end{equation}
where \({\text{SSIM}}_{k_{1}}^{\prime}\) denotes the \({\text{SSIM}}^{\prime}\) with
hyper-parameter \(k = k_{1}\) and \({\text{SSIM}}_{k_{2}}^{\prime}\) denotes the
\({\text{SSIM}}^{\prime}\) with hyper-parameter \(k = k_{2}\). If and only if
\(I_{x} = I_{y}\), we have
\({\text{SSIM}}^{\prime}( I_{x},\ I_{y} ) = \text{SSIM}( I_{x},\ I_{y} )\)
and
\({\text{SSIM}}_{k_{1}}^{\prime}( I_{x},\ I_{y} ) = {\text{SSIM}}_{k_{2}}^{\prime}( I_{x},\ I_{y} )\).
From Eq. (\ref{eq:14}), we find that our improved \({\text{SSIM}}^{\prime}\) usually obtain a
lower structural similarity index when compared to the original
\(\text{SSIM}\), thus increasing the percentage of structural similarity in the
overall loss function.

Following previous methods \cite{MD2,PLNet}, we compute per-pixel structural
similarity index with the corresponding local patch. As shown in Fig. \ref{fig:analysis}(a), the interval of valid structural similarity indexes is set to
{[}\(Lb\), \(Ub\){]}. The structural similarity indexes within the
interval {[}\(Ub\), 1{]} are too high to provide effective gradients for
depth estimation. Therefore, the low-texture regions with large amount of too high indexes make our network
difficult to train. The structural similarity index usually increases
gradually during training, indicating that the local patches corresponding to the two matched pixels become increasingly similar. However, some structural
similarity indexes in the noisy regions (e.g. occlusion area before
warping) tend to remain low values during training , which in turn cause the \text{SSIM} loss \(L_{\text{ssim}}\) to remain high values according to Eq. (\ref{eq:8}). Therefore, these low indexes that do not converge during training can be
detrimental to the optimization of the network. Here we use the interval
{[}-1, \(Lb\){]} to denote the range of indexes that are not conducive to the convergence of our models.

Fig. \ref{fig:analysis}(b) shows the percentage of invalid indexes within the interval
{[}\(Ub\), 1{]} in a sample of the same low-texture region. Since the
improved \({\text{SSIM}}^{\prime}\) can obtains a lower structural similarity index,
the percentage \(B\%\) in \({\text{SSIM}}^{\prime}\) is less than the percentage
\(A\%\) in \(\text{SSIM}\). Furthermore, it can be concluded from Eq. (\ref{eq:15}) that
the percentage \(B\%\) in \({\text{SSIM}}^{\prime}\) decreases as the
hyper-parameter \(k\) increases. Fig. \ref{fig:analysis}(c) shows the percentage of
invalid indexes within the interval {[}-1, \(Lb\){]} in a sample of the
same noisy region. We can easily conclude from Eq. (\ref{eq:14}) that the percentage
\(D\%\) in \({\text{SSIM}}^{\prime}\) is more than the percentage \(C\%\) in
\(\text{SSIM}\). Moreover, the percentage \(D\%\) in \({\text{SSIM}}^{\prime}\) increases
as the hyper-parameter \(k\) increases. In summary, for low-texture
regions, the larger \(k\) is the better, but for noisy regions, smaller
is better.

According to the above analysis, we find that \({\text{SSIM}}^{\prime}\) is more
sensitive to subtle differences. Specifically, for low-texture regions
with insignificant structural differences, the improved \({\text{SSIM}}^{\prime}\)
allows us to obtain more reasonable structural similarity indexes in contrast to the original very high and undiscriminating indexes. But for some
noisy regions, \({\text{SSIM}}^{\prime}\) generates more structural
similarity indexes which remain too low during training, thus severely affecting the optimization of the
network. It is worth mentioning that the percentage of low-texture
regions is usually much more than the noise regions in indoor images.
Therefore, the improved \({\text{SSIM}}^{\prime}\) can help us obtain more
reasonable structural similarity indexes over all pixels by setting a
suitable value for the hyper-parameter \(k\) to balance the effect in
low-texture regions and in noisy regions. Meanwhile, the decrease in
structural similarity index will in turn lead to an increase in
\(L_{{\text{ssim}}^{\prime}}\). The improved \(L_{{\text{ssim}}^{\prime}}\) here is defined as:
\begin{equation}
    L_{{\text{ssim}}^{\prime}} = \frac{\alpha}{2}( 1 - {\text{SSIM}}^{\prime}( I_{t},\ I_{t^{\prime} \rightarrow t} ) )
    \label{eq:16}
\end{equation}

So our novel photometric loss \(L_{{ph}^{\prime}}\) with an improved \text{SSIM}
function is computed as:
% \begin{small}
    \begin{equation}
    L_{{ph}^{\prime}}\!=\!\frac{\alpha}{2}( 1 - {\text{SSIM}}^{\prime}( I_{t},\ I_{t^{\prime} \rightarrow t} ) ) + (1 - \alpha)\| I_{t} - I_{t^{\prime} \rightarrow t}\|_{1}
\end{equation} 

% \end{small}
% \vspace{-5ex}
\subsection{Pose Network with Multiple Photometric Losses\label{sec:3C}}

\begin{figure*}[!htb]
    \centering
    \includegraphics[width=.9\textwidth]{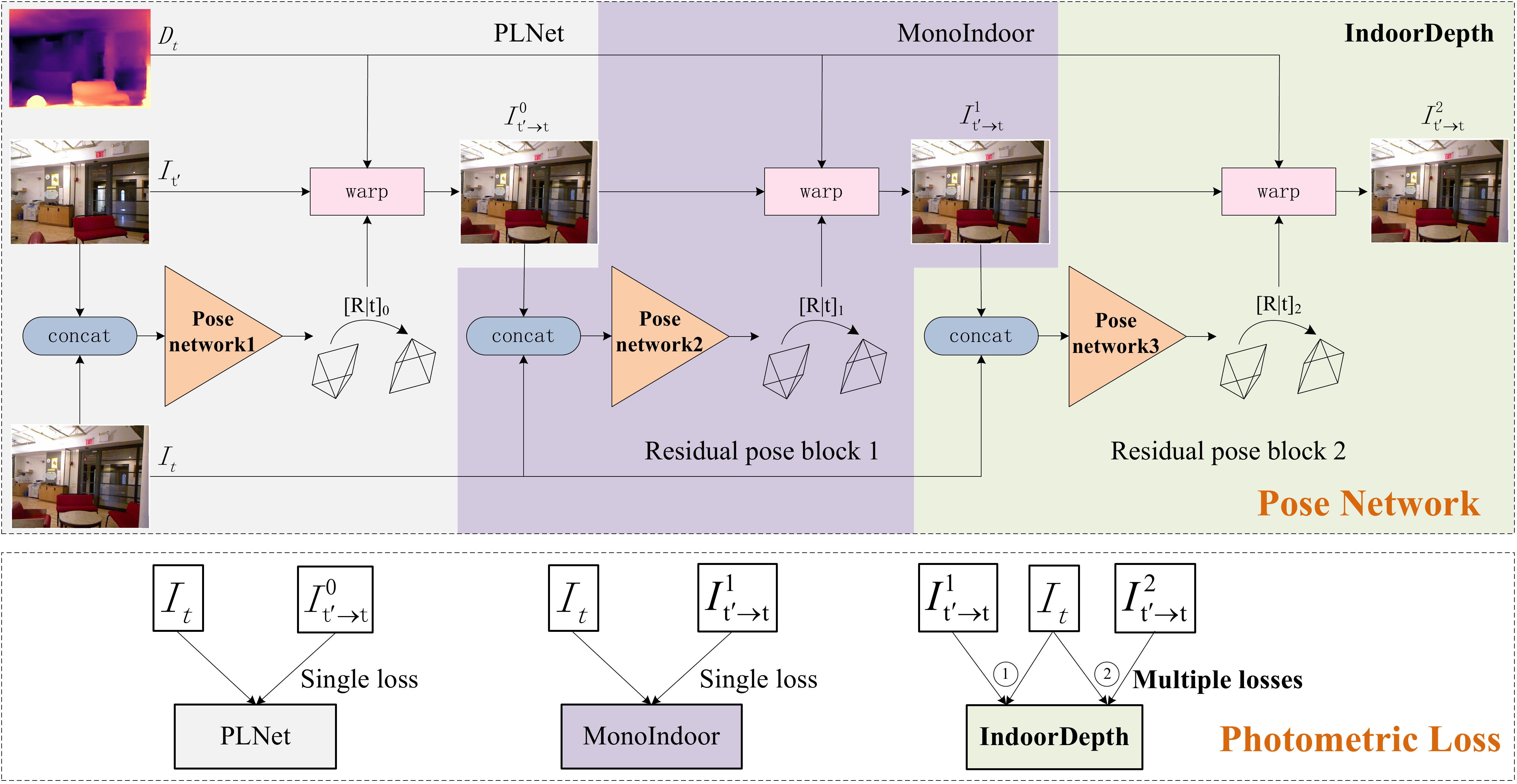}
    \caption{Proposed PoseNet with multiple photometric losses at
    different stages. The gray area shows the PoseNet of PLNet which has
    been described in detail in Section \ref{sec:3A}. The PoseNet of MonoIndoor
    consists of the gray area and the purple area. The PoseNet in our
    IndoorDepth, including all the areas, is a deeper network than previous
    approaches. Unlike the PLNet and MonoIndoor trained with only one
    photometric loss, we adopt multiple photometric losses to
    directly constrain both the intermediate residual pose block 1 and the
    last residual pose block 2.\label{fig:posnet}}
\end{figure*}

The main supervisory signal for self-supervised learning is the
photometric reprojection error, therefore it can be concluded from Eq.
(\ref{eq:2}) that both accurate depth and pose are essential. Most of the
previous methods employ a simple pose network, the same as the PLNet in
Fig. \ref{fig:posnet}. The results of these methods demonstrate that such a single pose network is
capable of estimating accurate poses in outdoor scenes (e.g. driving
scenes), since the camera poses are relative simple on outdoor
datasets. However, the ego-motion becomes more complicated in indoor
environments as the images are usually captured by hand-held devices.
Compared with camera poses in outdoor environments where translations
are dominant, indoor poses contain frequent rotations. This makes
accurate pose estimation on indoor datasets more difficult.

Following the iterative strategy in MonoIndoor \cite{MonoIndoor}, we design a
deeper pose network which can further alleviate the problem of inaccurate rotation prediction, as shown in Fig. \ref{fig:posnet}. Our PoseNet can be divided into three steps. In the
first step, the pose network takes the target image \(I_{t}\) and the
source image \(I_{t^{\prime}}\) as input and outputs the first relative pose
\(T_{t \rightarrow t^{\prime}}^{0}\). Then we can obtain the first
synthesized view \(I_{t^{\prime} \rightarrow t}^{0}\) by Eq. (\ref{eq:2}) and bilinear
interpolation. In the second step, the inputs change to the target image
\(I_{t}\) and synthesized view \(I_{t^{\prime} \rightarrow t}^{0}\). We can
subsequently get the second synthesized view
\(I_{t^{\prime} \rightarrow t}^{1}\) as the output of the second step. In the
last step, the third synthesized view \(I_{t^{\prime} \rightarrow t}^{2}\)
can be obtained by a similar process. Since pose networks output the
pose of the synthesized view relative to the target image in the latter
two steps, we call them residual pose blocks. It is worth mentioning
that all the pose networks shared the same pose encoder. But the
prediction heads of them learn independent parameters despite the same
structure.

As shown in Fig. \ref{fig:posnet}, there are two residual pose blocks in our PoseNet. However, further experiments in MonoIndoor show that the
performance does not significantly improve or even becomes worse when
the number of residual blocks is more than one. According to our conjecture, the reason for this failure may be that the PoseNet is too deep, leading
to a tendency to fall into local optima. In order to make all the
residual pose blocks work, we consider adding an additional photometric
loss to directly constrain the intermediate residual pose block, rather
than just constraining the last one. Specifically, multiple photometric
losses at different stages are used to train our PoseNet and DepthNet.
The two residual pose blocks in our PoseNet represent two different
stages. Here we use \(L_{ph}^{1}\) and \(L_{ph}^{2}\) to denote the
photometric losses in the two residual blocks, respectively. Thus they
can be calculated as:
\begin{align}
    L_{ph}^{1}\! &= \!\frac{\alpha}{2}( 1 - \text{SSIM}( I_{t}, I_{t^{\prime} \rightarrow t}^{1} ) ) + (1 - \alpha)\| I_{t} - I_{t^{\prime} \rightarrow t}^{1}\|_{1} \label{eq:18} \\
    L_{ph}^{2}\! &= \!\frac{\alpha}{2}( 1 - \text{SSIM}( I_{t}, I_{t^{\prime} \rightarrow t}^{2} ) ) + (1 - \alpha)\| I_{t} - I_{t^{\prime} \rightarrow t}^{2}\|_{1} \label{eq:19}
\end{align}
where \(I_{t^{\prime} \rightarrow t}^{1}\) and
\(I_{t^{\prime} \rightarrow t}^{2}\) denote the second synthesized view and
the third synthesized view, respectively. The final \(L_{ph}\) is a
weighted sum of each photometric loss, which is defined as:
\begin{equation}
    L_{ph} = L_{ph}^{1} + \beta L_{ph}^{2}
    \label{eq:20}
\end{equation}
\noindent where \(\beta\) is a hyper-parameter.

\begin{table}[!tb]
    \caption{Differences Between the PoseNet of MonoIndoor and That of IndoorDepth. "No. Residual Pose Block" Means the Number of Residual Pose Blocks in the PoseNet. Here \(\omega _{i}\) Denotes the Hyper-Parameter.\label{tab:multi-ph}}
    \centering
    % \small
    \renewcommand\arraystretch{1.7}
    \begin{tabularx}{.8\linewidth}{cCC}
        % \hline
        \toprule
        Methods & No. Residual Pose Block & Photometric Loss \\ 
        \midrule
        MonoIndoor \cite{MonoIndoor} & \(n_{1}\) &
        \(L_{ph}^{n_{1}}\) \\ %\hline
        IndoorDepth (ours) & \(n_{2}\) &
        \(\sum_{i = 1}^{n_{2}} \omega _{i} L_{ph}^{i}\) \\
% \hline
        \bottomrule
    \end{tabularx}
\end{table}
Table \ref{tab:multi-ph} shows the essential differences between the PoseNet of MonoIndoor \cite{MonoIndoor} and that of our IndoorDepth. In fact, each method can construct a pose network with an arbitrary number of residual pose blocks. Here we use \(n_{1}\) and \(n_{2}\) to denote the number of residual pose blocks in MonoIndoor and IndoorDepth, respectively. The novelty of our approach is that we use multiple photometric losses at different stages to train the pose network, while MonoIndoor only use the photometric loss \(L_{ph}^{n_{1}}\) at the last stage. Further experiments in MonoIndoor \cite{MonoIndoor} demonstrate that the best performance can be obtained when \(n_{1} = 1\), which is consistent with the description of MonoIndoor in Fig. \ref{fig:posnet}. However, experiments in subsequent sections show that our IndoorDepth can obtain the best performance when \(n_{2} = 2\) and \(\omega_{1} = \omega_{2} = 1 \) and the performance is superior to that in MonoIndoor.
% \vspace{-3ex}
\subsection{Overall Loss Function\label{sec:3D}}
We use our novel photometric loss \(L_{{ph}^{\prime}}\) instead of the
original photometric loss \(L_{ph}\) and employ multiple photometric
losses as defined in Eq. (\ref{eq:20}). Based on the baseline loss in Eq. (\ref{eq:7}), our overall loss function is defined as:
\begin{equation}
    L = L_{{ph}^{\prime}}^{1} + \beta L_{{ph}^{\prime}}^{2} + {\alpha_{sm}L}_{sm} + {\alpha_{pc}L}_{pc} + {\alpha_{lc}L}_{lc}
    \label{eq:21}
\end{equation}
where \(L_{{ph}^{\prime}}^{1}\) and \(L_{{ph}^{\prime}}^{2}\) are obtained by
using \({\text{SSIM}}^{\prime}\) instead of \(\text{SSIM}\) in Eq. (\ref{eq:18}) and Eq. (\ref{eq:19}),
respectively. \(\beta\), \(\alpha_{sm}\), \(\alpha_{pc}\) and
\(\alpha_{lc}\) are hyper-parameters.

\section{Experiments}
\subsection{Implementation Details}
In this paper, we use PLNet \cite{PLNet} as the backbone model for depth
prediction and implement our model using the PyTorch framework. For the
hyper-parameters in Eq. (\ref{eq:21}), we set \(\beta = 1.0\),
\(\alpha_{sm} = 0.2\), \(\alpha_{pc} = 2.0\) and \(\alpha_{lc} = 0.5\).
We use \(k = 5.0\) in Eq. (\ref{eq:11}) for our novel photometric loss. We
jointly train the DepthNet and PoseNet for 30 epochs with a batch size
of 16 using the Adam \cite{Adam} optimizer. The initial learning rate is set
to \(10^{- 4}\) for the first 20 epochs and then it is decayed once by
0.1 for the last 10 epochs. Our model is trained with 3 frames,
including a target view and two source views, on a single Nvidia 2080Ti
GPU. The training time of our IndoorDepth is about 9 hours.

\begin{table*}[!h!tb]
    \caption{Quantitative Comparison of Our IndoorDepth to Other Existing Methods on
    NYUv2. Best Results in Each Category are in Bold. For Error Evaluating
    Metrics, Lower is Better. For Accuracy Evaluating Metrics, Higher is Better.\label{tab:NYUv2}}
    \centering
  \renewcommand\arraystretch{1.2}
    \begin{tabularx}{.85\textwidth}{lcCCCCCC}
    % \begin{tabularx}{.8\textwidth}{|c|YZC|C|C|C|Y}
    % \hhline{|*{8}-|}
    \toprule
    % \hline
                            &                               & \multicolumn{3}{c}{Error Metric ↓ }                                                                              & \multicolumn{3}{c}{Accuracy Metric ↑}                                                   \\ \cmidrule(r){3-5} \cmidrule(r){6-8}%\hhline{|~|~||*{6}{-|}}%\cline{3-8}%\specialrule{1pt}{0pt}{0pt}%\cline{3-8} % 
    \multirow{-2}{*}[2pt]{Methods} & \multirow{-2}{*}[2pt]{Supervision} & Abs Rel & Log10 & RMS & $\delta$ \textless 1.25 & $\delta$ \textless 1.25$^2$ & $\delta$ \textless 1.25$^3$ \\ %\hhline{|*{2}=::*{6}=|}
    % \hhline{|-|-||*{5}{-|}-|}
    \midrule
    % \hline
    % \cline{1-2}
    Depth Transfer \cite{DepthTransfer}                                                   & D                                                                            & 0.349                           & 0.131                         & 1.21                        & -                                        & -                                         & -                                         \\
    Li et al. \cite{Li}                                                        & D                                                                            & 0.232                           & 0.094                         & 0.821                       & 0.621                                    & 0.886                                     & 0.968                                     \\
    Wang et al. \cite{Wang}                                                     & D                                                                            & 0.220                           & 0.094                         & 0.745                       & 0.605                                    & 0.890                                     & 0.970                                     \\
    Eigen et al. \cite{Eigen1}                                                    & D                                                                            & 0.215                           & -                             & 0.907                       & 0.611                                    & 0.887                                     & 0.971                                     \\
    Liu et al. \cite{Supervised6}                                                      & D                                                                            & 0.213                           & 0.087                         & 0.759                       & 0.650                                    & 0.906                                     & 0.976                                     \\
    Eigen et al. \cite{Supervised1}                                                     & D                                                                            & 0.158                           & -                             & 0.641                       & 0.769                                    & 0.950                                     & 0.988                                     \\
    Chakrabarti et al. \cite{Supervised2}                                               & D                                                                            & 0.149                           & -                             & 0.620                       & 0.806                                    & 0.958                                     & 0.987                                     \\
    Li et al. \cite{Supervised4}                                                        & D                                                                            & 0.143                           & 0.063                         & 0.635                       & 0.788                                    & 0.958                                     & 0.991                                     \\
    FCRN \cite{Supervised3}                                                             & D                                                                            & 0.127                           & 0.055                         & 0.573                       & 0.811                                    & 0.953                                     & 0.988                                     \\
    Xu et al. \cite{Supervised7}                                                        & D                                                                            & 0.121                           & 0.052                         & 0.586                       & 0.811                                    & 0.954                                     & 0.987                                     \\
    DORN \cite{DORN}                                                             & D                                                                            & 0.115                           & 0.051                         & 0.509                       & 0.828                                    & 0.965                                     & 0.992                                     \\
    Hu et al. \cite{Supervised9}                                                        & D                                                                            & 0.115                           & 0.050                         & 0.530                       & 0.866                                    & 0.975                                     & 0.993                                     \\
    VNL \cite{Supervised10}                                                             & D                                                                            & 0.108                           & \textbf{0.048}                & 0.416                       & \textbf{0.875}                           & \textbf{0.976}                            & \textbf{0.994}                            \\
    Fang et al. \cite{Fang}                                                  & D                                                                            & \textbf{0.101}                  & -                             & \textbf{0.412}              & 0.868                                    & 0.958                                     & 0.986                                     \\
    % \hhline{|-|-||*{6}{-|}}
    % \hline
    \midrule
    DistDepth \cite{DistDepth}                                                        & MS                                                                           & 0.130                           & -                             & 0.517                       & 0.832                                    & 0.963                                     & 0.990                                     \\
    % \hhline{|-|-||*{6}{-|}}
    \midrule
    % \hline
    MovingIndoor \cite{MovingIndoor}                                                     & M                                                                            & 0.208                           & 0.086                         & 0.712                       & 0.674                                    & 0.900                                     & 0.968                                     \\
    TrainFlow \cite{TrainFlow}                                                        & M                                                                            & 0.189                           & 0.079                         & 0.686                       & 0.701                                    & 0.912                                     & 0.978                                     \\
    Monodepth2 \cite{MD2}                                                       & M                                                                            & 0.171                           & 0.071                         & 0.639                       & 0.750                                    & 0.945                                     & 0.987                                     \\
    P\textsuperscript{2}Net (3 frames) \cite{P2Net}                                                 & M                                                                            & 0.159                           & 0.068                         & 0.599                       & 0.772                                    & 0.942                                     & 0.984                                     \\
    P\textsuperscript{2}Net (5 frames) \cite{P2Net}                                                 & M                                                                            & 0.150                           & 0.064                         & 0.561                       & 0.796                                    & 0.948                                     & 0.986                                     \\
    PLNet (3 frames) \cite{PLNet}                                                 & M                                                                            & 0.151                           & 0.064                         & 0.562                       & 0.790                                    & 0.953                                     & 0.989                                     \\
    PLNet (5 frames) \cite{PLNet}                                                 & M                                                                            & 0.144                           & 0.061                         & 0.540                       & 0.807                                    & 0.957                                     & 0.990                                     \\
    IFMNet \cite{IFMNet}                                                 & M                                                                            & 0.145                           & 0.063                         & 0.569                       & 0.799                                    & 0.949                                     & 0.986                                     \\
    StructDepth \cite{StructDepth}                                                      & M                                                                            & 0.142                           & 0.060                         & 0.540                       & 0.813                                    & 0.954                                     & 0.988                                     \\
    ARN \cite{ARN}                                                              & M                                                                            & 0.138                           & 0.059                         & 0.532                       & 0.820                                    & 0.956                                     & 0.989                                     \\
    MonoIndoor \cite{MonoIndoor}                                                       & M                                                                            & 0.134                           & -                             & 0.526                       & 0.823                                    & 0.958                                     & 0.989                                     \\
    MonoIndoor++ \cite{MonoIndoor++}                                                       & M                                                                            & 0.132                           & -                             & 0.517                       & 0.834                                    & 0.961                                     & 0.990                                     \\
    \textbf{IndoorDepth (ours)}                                              & M                                                                            & \textbf{0.126}                  & \textbf{0.054}                & \textbf{0.494}              & \textbf{0.845}                           & \textbf{0.965}                            & \textbf{0.991}\\                           
    % \hline
        % \thickhline
        \bottomrule
    \end{tabularx}
\end{table*}

\begin{figure*}[!h!tb]
    \centering
    \vspace{-13pt}
    \begin{tabularx}{.96\textwidth}{CCCCC}
        Input\vspace{-12pt} & PLNet \cite{PLNet}\vspace{-12pt} & StructDepth \cite{StructDepth}\vspace{-12pt}& Ours\vspace{-12pt} & GT 
        \vspace{-12pt}
    \end{tabularx}
    % \vspace{-4pt}
    \subfloat{
        \includegraphics[width=.19\textwidth]{4-1-1.png}
    }
    \hspace{-9pt}
    \subfloat{
        \includegraphics[width=.19\textwidth]{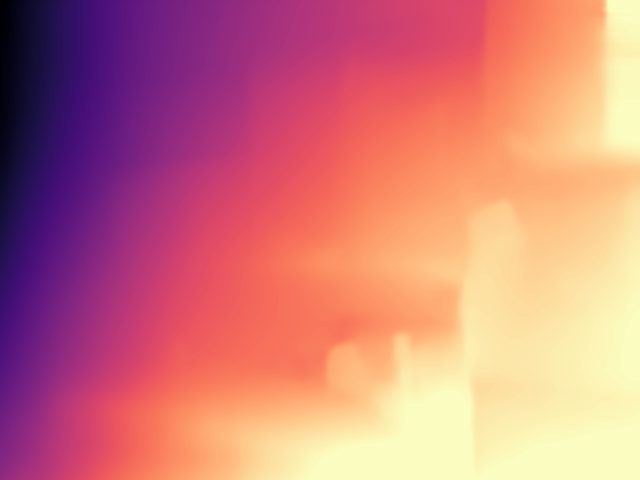}
    }
    \hspace{-8pt}
    \subfloat{
        \includegraphics[width=.19\textwidth]{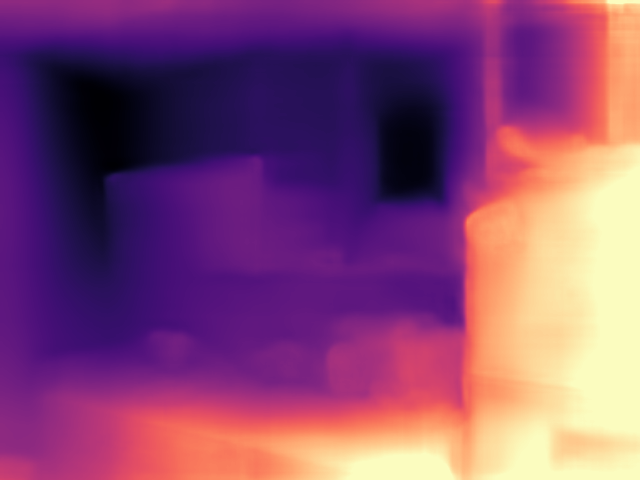}
    }\hspace{-8pt}
    \subfloat{
        \includegraphics[width=.19\textwidth]{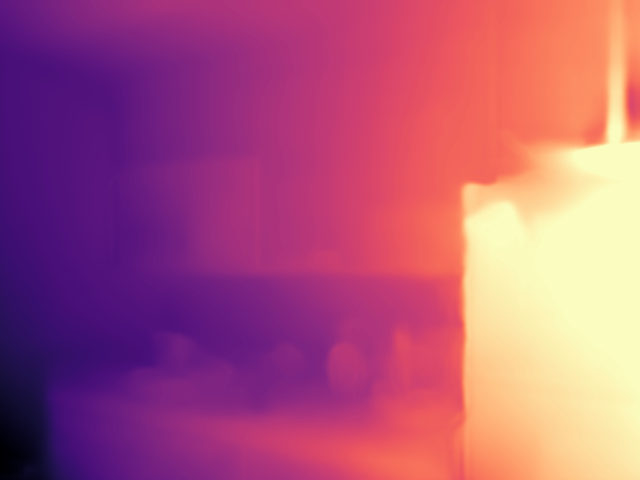}
    }\hspace{-8pt}
    \subfloat{
        \includegraphics[width=.19\textwidth]{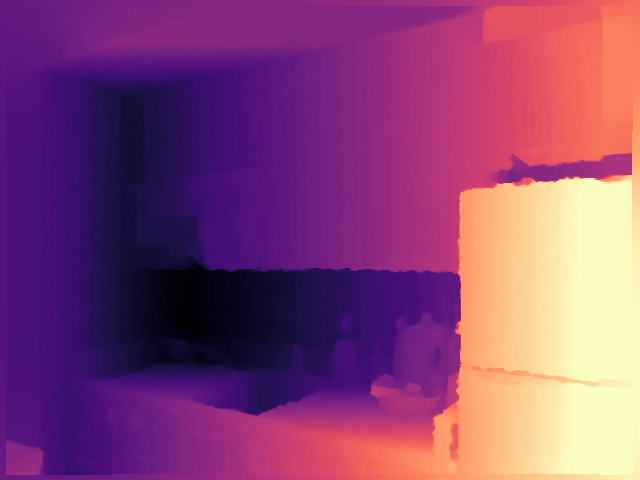}
    }
    \\ \vspace{-8pt}
    \subfloat{
        \includegraphics[width=.19\textwidth]{4-2-1.png}
    }\hspace{-9pt}
    \subfloat{
        \includegraphics[width=.19\textwidth]{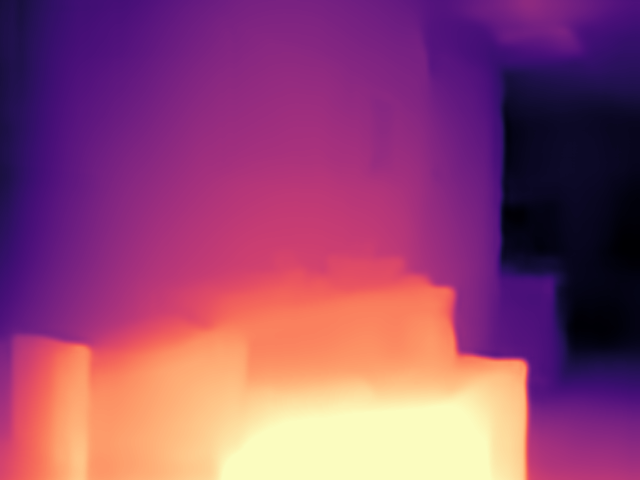}
    }\hspace{-8pt}
    \subfloat{
        \includegraphics[width=.19\textwidth]{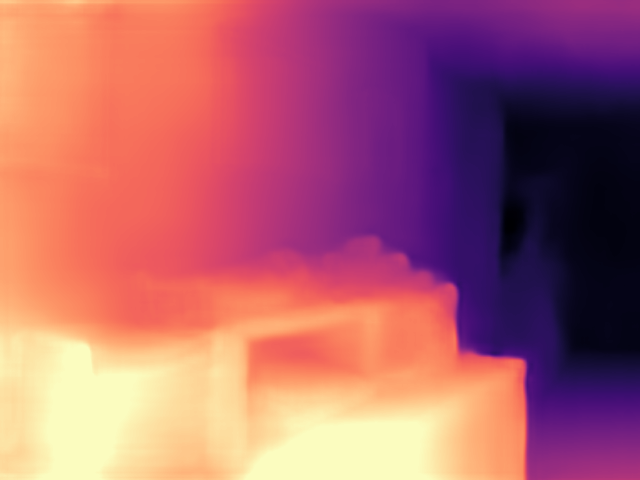}
    }\hspace{-8pt}
    \subfloat{
        \includegraphics[width=.19\textwidth]{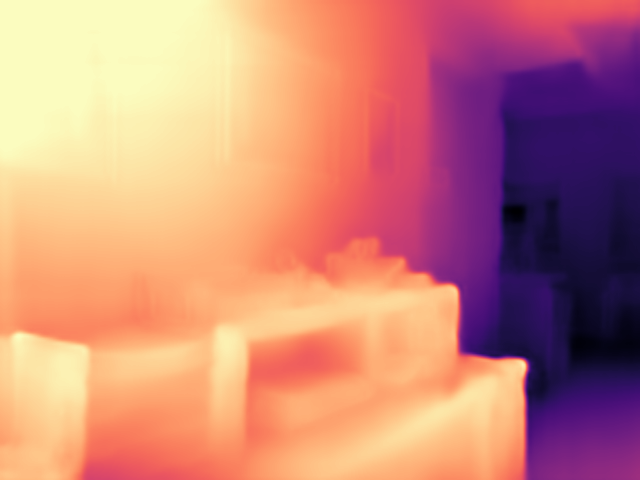}
    }\hspace{-8pt}
    \subfloat{
        \includegraphics[width=.19\textwidth]{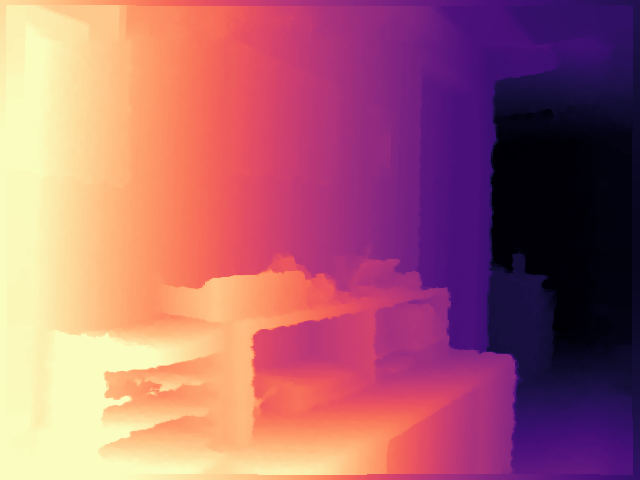}
    }
    \\ \vspace{-8pt}
    \subfloat{
        \includegraphics[width=.19\textwidth]{4-3-1.png}
    }\hspace{-9pt}
    \subfloat{
        \includegraphics[width=.19\textwidth]{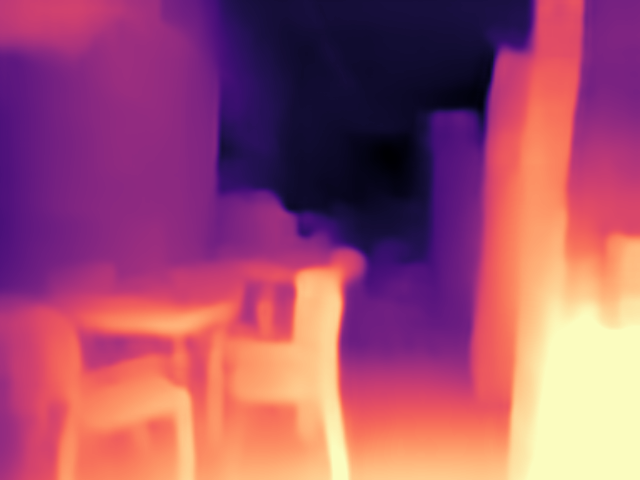}
    }\hspace{-8pt}
    \subfloat{
        \includegraphics[width=.19\textwidth]{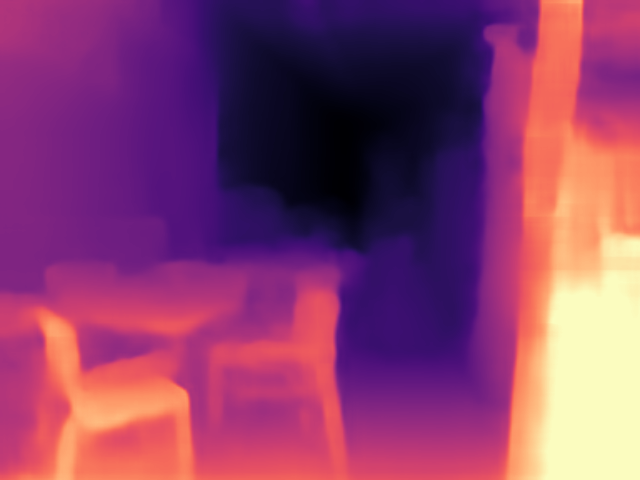}
    }\hspace{-8pt}
    \subfloat{
        \includegraphics[width=.19\textwidth]{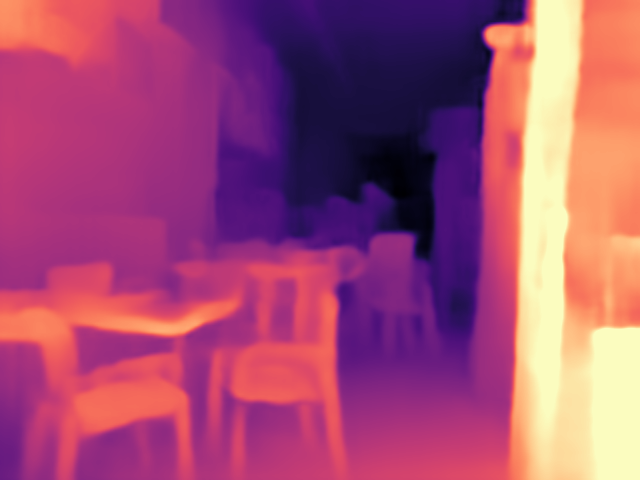}
    }\hspace{-8pt}
    \subfloat{
        \includegraphics[width=.19\textwidth]{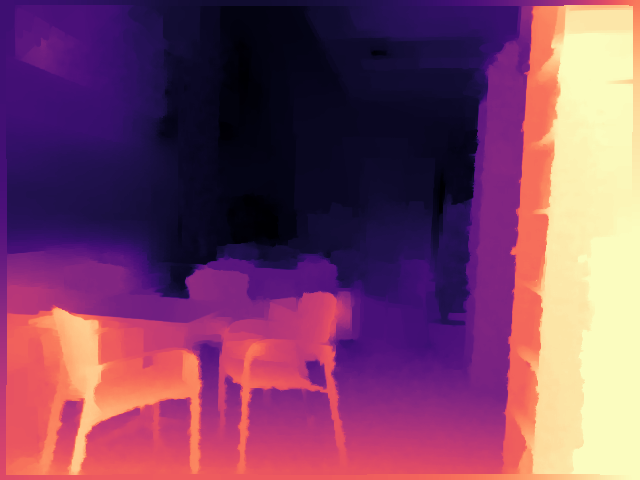}
    }
    \\ \vspace{-8pt}
    \subfloat{
        \includegraphics[width=.19\textwidth]{4-4-1.png}
    }\hspace{-9pt}
    \subfloat{
        \includegraphics[width=.19\textwidth]{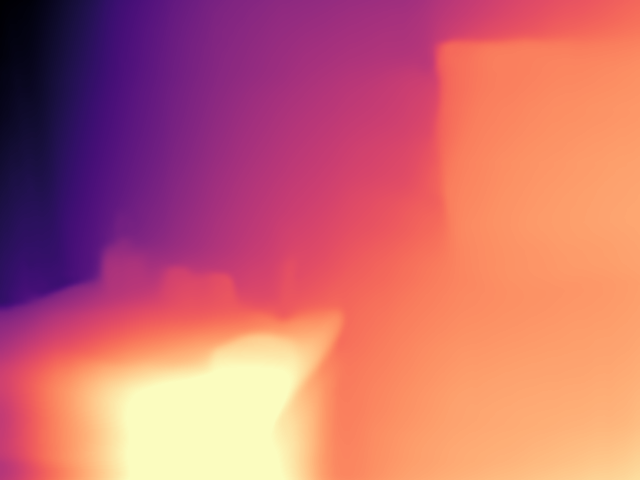}
    }\hspace{-8pt}
    \subfloat{
        \includegraphics[width=.19\textwidth]{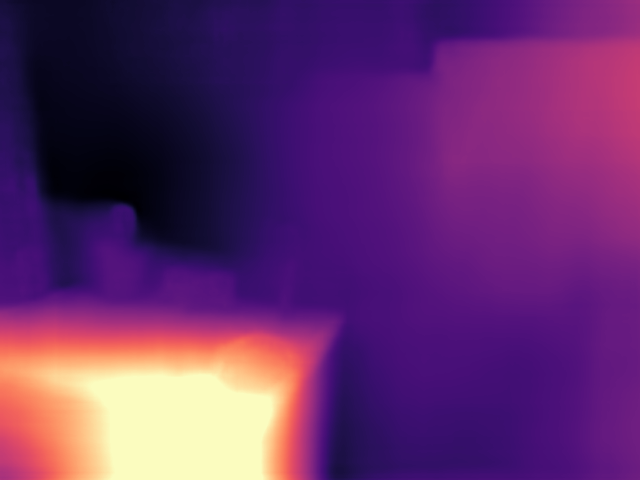}
    }\hspace{-8pt}
    \subfloat{
        \includegraphics[width=.19\textwidth]{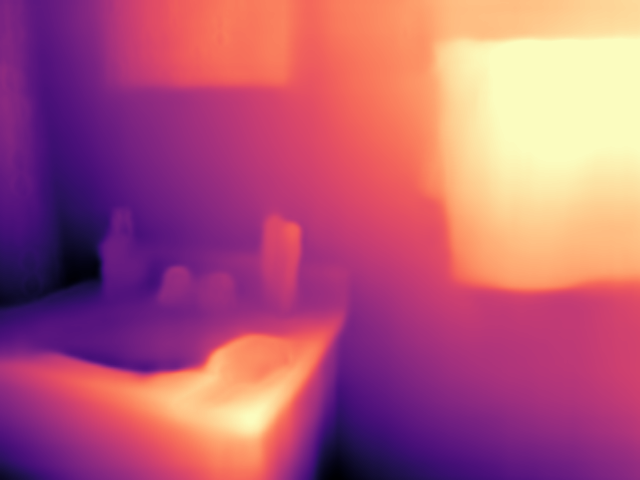}
    }\hspace{-8pt}
    \subfloat{
        \includegraphics[width=.19\textwidth]{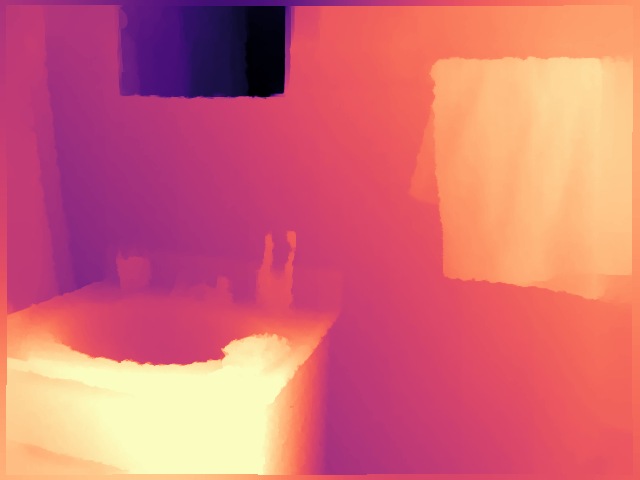}
    }\hspace{-8pt}
    \\ \vspace{-8pt}
    \subfloat{
        \includegraphics[width=.19\textwidth]{4-5-1.png}
    }\hspace{-9pt}
    \subfloat{
        \includegraphics[width=.19\textwidth]{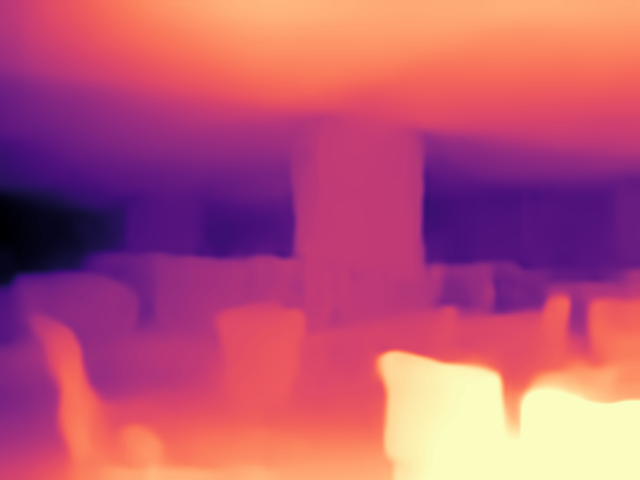}
    }\hspace{-8pt}
    \subfloat{
        \includegraphics[width=.19\textwidth]{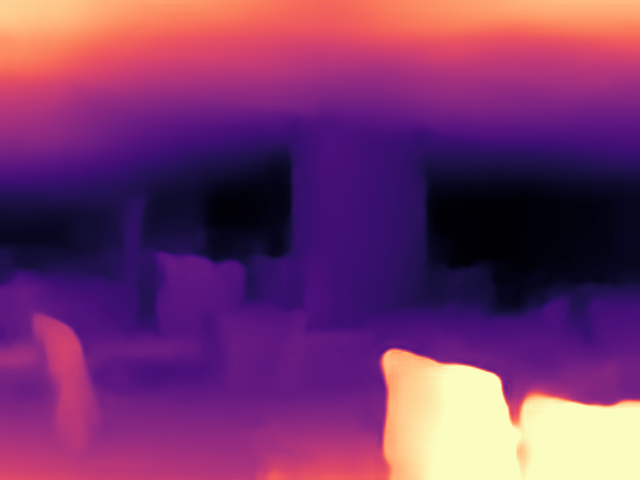}
    }\hspace{-8pt}
    \subfloat{
        \includegraphics[width=.19\textwidth]{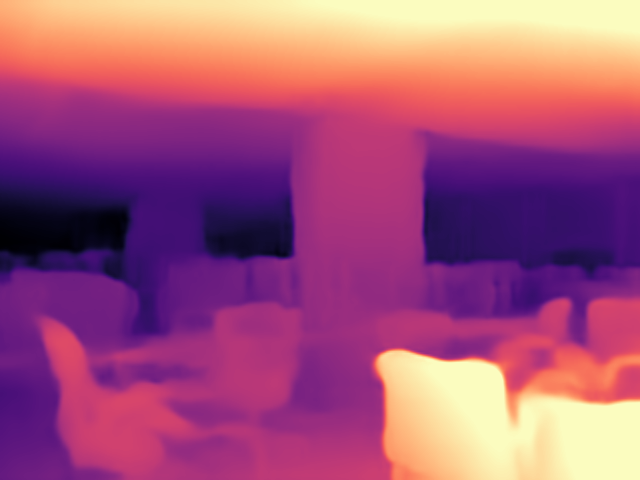}
    }\hspace{-8pt}
    \subfloat{
        \includegraphics[width=.19\textwidth]{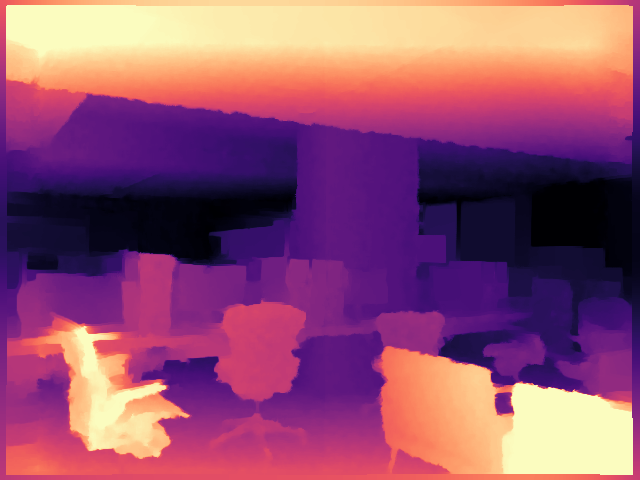}
    }\hspace{-8pt}
    \\ \vspace{-8pt}
    \subfloat{
        \includegraphics[width=.19\textwidth]{4-6-1.png}
    }\hspace{-9pt}
    \subfloat{
        \includegraphics[width=.19\textwidth]{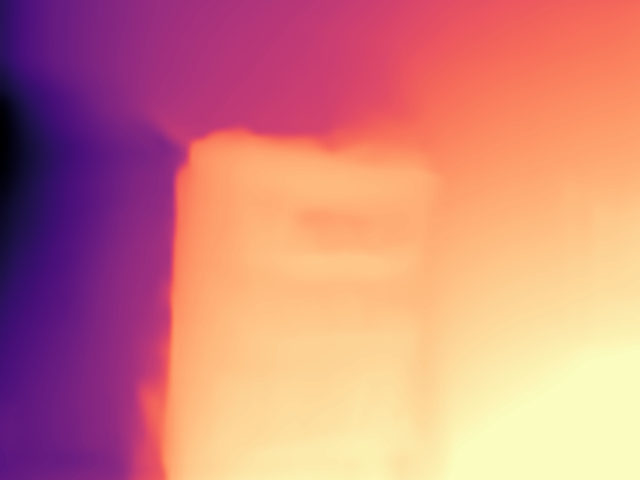}
    }\hspace{-8pt}
    \subfloat{
        \includegraphics[width=.19\textwidth]{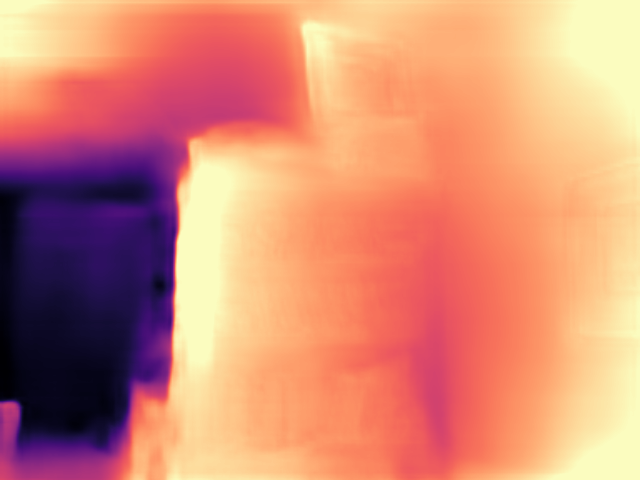}
    }\hspace{-8pt}
    \subfloat{
        \includegraphics[width=.19\textwidth]{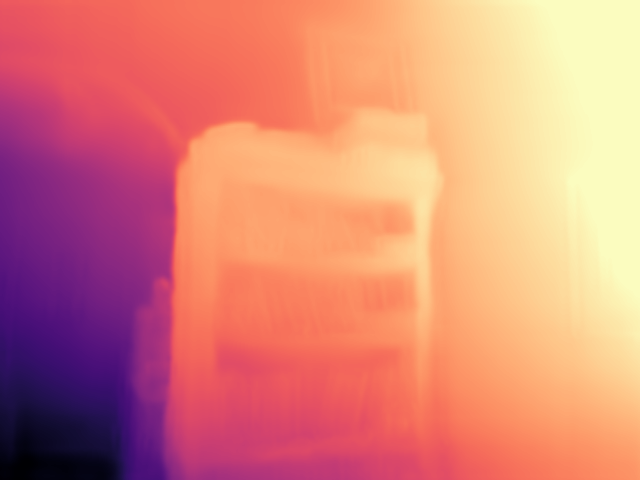}
    }\hspace{-8pt}
    \subfloat{
        \includegraphics[width=.19\textwidth]{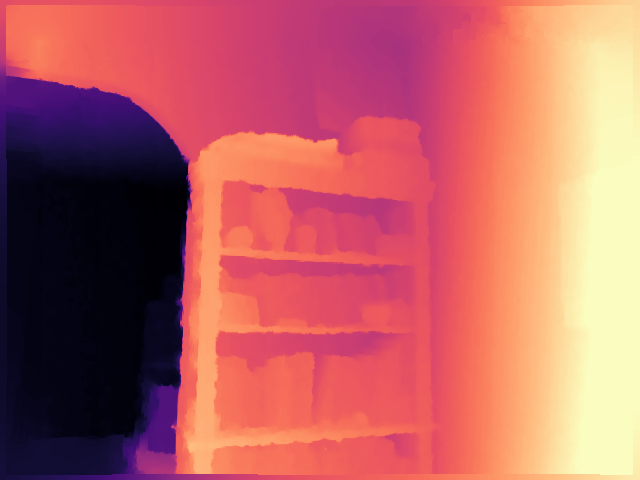}
    }\hspace{-8pt}\\
    \caption{Qualitative results on NYUv2 dataset \cite{NYUv2}. Compared with PLNet
    \cite{PLNet} and StructDepth \cite{StructDepth}, our model, IndoorDepth, produces the
    sharpest depth maps that are closest to the ground-truth depth. Better
    viewed by zooming on screen.\label{fig:NYUv2}}
\end{figure*}
% \vspace{-10pt}
\subsection{Dataset and Metrics}
\subsubsection{NYUv2 \cite{NYUv2}}
NYUv2 is the most popular dataset for indoor depth
estimation. We train the proposed IndoorDepth model on the NYUv2 dataset and use the official 654 images with ground-truth depth for
testing. Following \cite{PLNet}, we first sample the raw dataset at 5 frames
interval as our target images and obtain about 47K images for training
as a result. In order to remove redundant images and obtain enough
translation, we sample the sequences by \(\pm\)10 frames as our source
images. Then we follow previous works \cite{P2Net,DeepV2D} to undistort all
images and crop 16 pixels from the border regions. Finally, the images
are resized to \(320 \times 256\) for training. 
% \newpage

\subsubsection{ScanNet \cite{ScanNet}} ScanNet is an RGB-D video dataset including around
2.5M images in more than 1500 scenes. In this paper, we directly
evaluate our pre-trained model on NYUv2 dataset to validate the
generalization ability of our method. We use the same test split as in
previous methods \cite{P2Net,PLNet,StructDepth} including 533 images, which are
randomly selected from various scenes. The test images are also resized
to \(320 \times 256\) pixels.

\subsubsection{Metrics} Following \cite{Supervised3,Supervised6,DORN,Supervised10}, standard metrics
including three error evaluating metrics and three accuracy evaluating
metrics are used to evaluate our depth prediction. Specifically, error
evaluating metrics consist of the mean absolute relative error (AbsRel),
mean log10 error (Log10) and root mean squared error (RMS), which are
defined as:
\begin{align}
AbsRel &= \frac{1}{T}\sum_{i = 1}^{T}\frac{\left| d_{i} - d_{i}^{*} \right|}{d_{i}^{*}} \\
Log10 &= \frac{1}{T}\sum_{i = 1}^{T}\left| lg(d_i) - lg(d_{i}^{*}) \right| \\
RMS &= \sqrt{\frac{1}{T}\sum_{i = 1}^{T}\left\| d_{i} - d_{i}^{*} \right\|^{2}}
\end{align}
where \(d_{i}\) and \(d_{i}^{*}\) denote the predicted depth and
ground-truth depth at the pixel indexed by \(i\), respectively. \(T\) is
the sum of the number of pixels in all test images. For accuracy
evaluating metrics, we calculate the accuracy values under three
different thresholds by the following equation:
\begin{equation}
    max(\frac{d_{i}}{d_{i}^{*}},\ \frac{d_{i}^{*}}{d_{i}}) = \delta < Thr
\end{equation}
where \(Thr = {1.25}^{t}\) (\(t \in \{ 1,2,3\}\)) denotes the threshold.
As self-supervised methods based on monocular sequences cannot obtain
the metric scale, we use the same median scaling strategy as that in
\cite{MD2,SC-Depth} for testing. The depth is capped at 10m in both NYUv2 and
ScanNet.

\subsection{NYUv2 Results}
\subsubsection{Quantitative Results}
Table \ref{tab:NYUv2} shows the monocular depth estimation
results on NYUv2 \cite{NYUv2}. All the methods can be divided into three
categories according to different supervisions. D denotes depth
supervision, M denotes monocular video supervision and MS means
monocular video combined with stereo pairs. The results show that our
IndoorDepth outperforms previous self-supervised models of M by
a large margin and even beats the model of MS. Specifically, compared to
the state-of-the-art self-supervised approach, MonoIndoor++ \cite{MonoIndoor++}, our model
reduces AbsRel and RMS by 0.6\% and 2.3\%, respectively. The three
accuracy evaluating metrics are increased by 1.1\%, 0.4\% and 0.1\%,
respectively. In addition, our method outperforms a large number of
supervised methods and further closes the performance gap between
supervised methods and self-supervised methods.

\subsubsection{Qualitative Results}
In Fig. \ref{fig:NYUv2} we present qualitative results for our
model on NYUv2 dataset \cite{NYUv2}. We can see that our model generates
higher quality depth maps when compared with PLNet \cite{PLNet} and
StructDepth \cite{StructDepth}. To be specific, our depth maps have sharper edges
than others, such as the chairs in the fifth row and the bookshelf in
the last row. Furthermore, we observe that depth predictions in
low-texture regions are closer to the ground-truth results, such as the
walls in the second and fourth rows. The results demonstrate that our
model predicts more accurate depth in both object boundaries and
low-texture regions.

\subsection{ScanNet Results}
\subsubsection{Quantitative Results}
Table \ref{tab:scannet} shows the quantitative results on
ScanNet \cite{ScanNet}, where all models are trained only on NYUv2. The
evaluation results show that our method outperforms previous
state-of-the-art self-supervised methods on ScanNet. Compared with StructDepth
\cite{StructDepth}, it is obviously that our approach has better performance across
all metrics. Moreover, our method reduces the gap between
supervised methods and self-supervised methods.

\begin{table*}[!t!b]
    \caption{Quantitative Comparison on Generalization. All Models are Trained Only
    on NYUv2 and Evaluated on ScanNet. Best Results in Each Category are in
    Bold. For Error Evaluating Metrics, Lower is Better. For
    Accuracy Evaluating Metrics, Higher is Better.\label{tab:scannet}}
    \centering
    \renewcommand\arraystretch{1.2}
    \begin{tabularx}{.85\textwidth}{lc*{6}{C}}
    
    \toprule
      &   & \multicolumn{3}{c}{Error Metric ↓}  & \multicolumn{3}{c}{Accuracy Metric ↑}     \\ \cmidrule(r){3-5} \cmidrule(r){6-8}
    \multirow{-2}{*}[2pt]{Methods} & \multirow{-2}{*}[2pt]{Supervision} & Abs Rel & Log10 & RMS & $\delta$ \textless 1.25 & $\delta$ \textless 1.25$^2$ & $\delta$ \textless 1.25$^3$ \\ %\hhline{|*{2}=::*{6}=|}
    \midrule
    % \thickhline
    FCRN \cite{Supervised3} & D & 0.141 & 0.059 & 0.339 & 0.811 & 0.958 & 0.990 \\
VNL \cite{Supervised10} & D & \textbf{0.123} & \textbf{0.052} & \textbf{0.306} &
\textbf{0.848} & \textbf{0.964} & \textbf{0.991} \\
\midrule
MovingIndoor \cite{MovingIndoor} & M & 0.212 & 0.088 & 0.483 & 0.650 & 0.905 &
0.976 \\
Monodepth2 \cite{MD2} & M & 0.200 & 0.083 & 0.458 & 0.672 & 0.922 &
0.981 \\
TrainFlow \cite{TrainFlow} & M & 0.179 & 0.076 & 0.415 & 0.726 & 0.927 & 0.980 \\
P\textsuperscript{2}Net \cite{P2Net} & M & 0.175 & 0.074 & 0.420 & 0.740 &
0.932 & 0.982 \\
PLNet (3 frames) \cite{PLNet} & M & 0.176 & 0.074 & 0.414 & 0.735 & 0.939 &
0.985 \\
PLNet (5 frames) \cite{PLNet} & M & 0.168 & 0.072 & 0.404 & 0.750 & 0.942 &
0.985 \\
IFMNet \cite{IFMNet} & M & 0.170 & 0.071 & 0.402 & 0.758 & 0.940 &
0.989 \\
SC-Depth \cite{SC-Depth} & M & 0.169 & 0.072 & 0.392 & 0.749 & 0.938 & 0.983 \\
StructDepth \cite{StructDepth} & M & 0.165 & 0.070 & 0.400 & 0.754 & 0.939 &
0.985 \\
\textbf{IndoorDepth} & M & \textbf{0.153} & \textbf{0.065} & \textbf{0.373} &
\textbf{0.786} & \textbf{0.950} & \textbf{0.988} \\
    \bottomrule
    \end{tabularx}

\end{table*}

\subsubsection{Qualitative Results} 
\begin{figure*}[!t!b]
    % \vspace{-12pt}
    \centering
    \begin{tabularx}{.82\textwidth}{CCCCC}
        Input\vspace{-12pt} & PLNet \cite{PLNet}\vspace{-12pt} & StructDepth \cite{StructDepth}\vspace{-12pt} & Ours\vspace{-12pt} & GT\vspace{-12pt} \\
    \end{tabularx}\\
    \subfloat{
        \includegraphics[width=.16\textwidth]{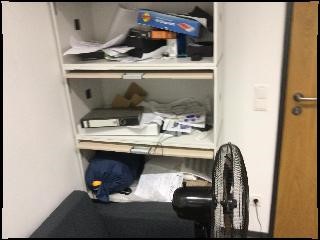}
    }
    \hspace{-9pt}
    \subfloat{
        \includegraphics[width=.16\textwidth]{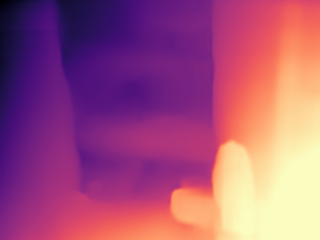}
    }
    \hspace{-8pt}
    \subfloat{
        \includegraphics[width=.16\textwidth]{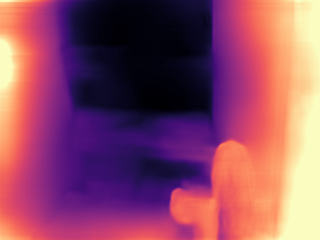}
    }\hspace{-8pt}
    \subfloat{
        \includegraphics[width=.16\textwidth]{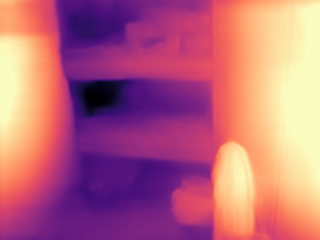}
    }\hspace{-8pt}
    \subfloat{
        \includegraphics[width=.16\textwidth]{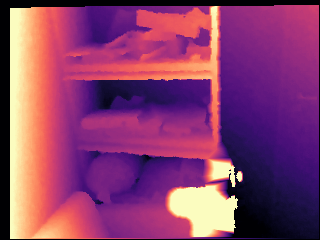}
    }
    \\ \vspace{-8pt}
    \subfloat{
        \includegraphics[width=.16\textwidth]{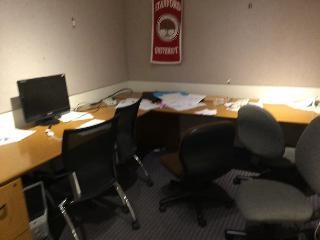}
    }\hspace{-9pt}
    \subfloat{
        \includegraphics[width=.16\textwidth]{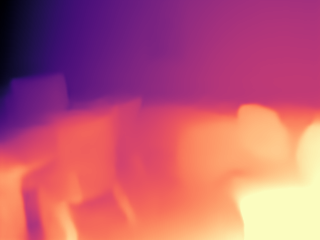}
    }\hspace{-8pt}
    \subfloat{
        \includegraphics[width=.16\textwidth]{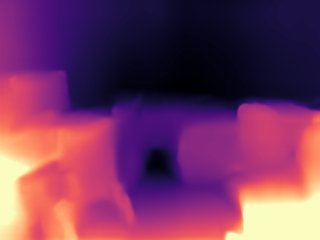}
    }\hspace{-8pt}
    \subfloat{
        \includegraphics[width=.16\textwidth]{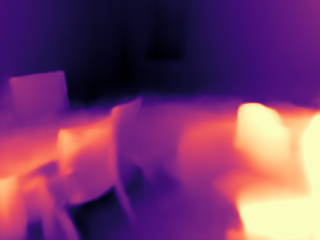}
    }\hspace{-8pt}
    \subfloat{
        \includegraphics[width=.16\textwidth]{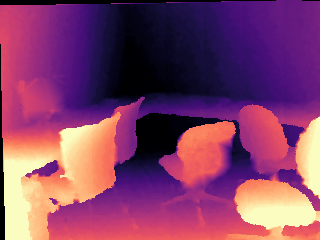}
    }
    \\ \vspace{-8pt}
    \subfloat{
        \includegraphics[width=.16\textwidth]{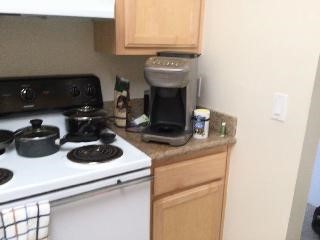}
    }\hspace{-9pt}
    \subfloat{
        \includegraphics[width=.16\textwidth]{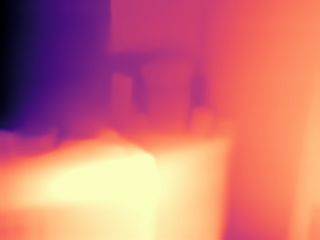}
    }\hspace{-8pt}
    \subfloat{
        \includegraphics[width=.16\textwidth]{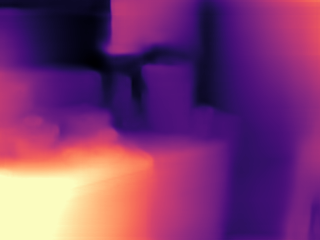}
    }\hspace{-8pt}
    \subfloat{
        \includegraphics[width=.16\textwidth]{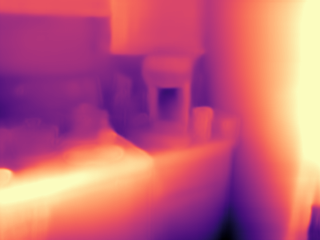}
    }\hspace{-8pt}
    \subfloat{
        \includegraphics[width=.16\textwidth]{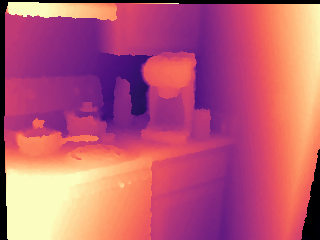}
    }
    \\ \vspace{-8pt}
    \subfloat{
        \includegraphics[width=.16\textwidth]{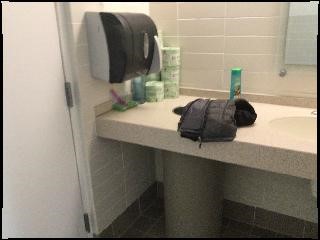}
    }\hspace{-9pt}
    \subfloat{
        \includegraphics[width=.16\textwidth]{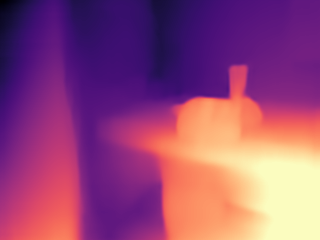}
    }\hspace{-8pt}
    \subfloat{
        \includegraphics[width=.16\textwidth]{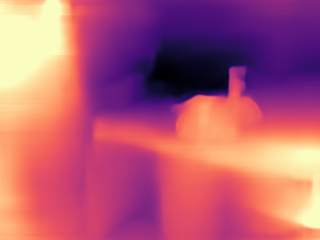}
    }\hspace{-8pt}
    \subfloat{
        \includegraphics[width=.16\textwidth]{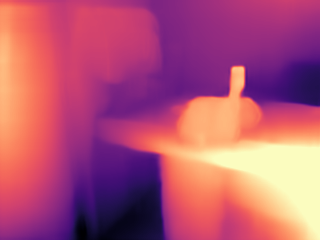}
    }\hspace{-8pt}
    \subfloat{
        \includegraphics[width=.16\textwidth]{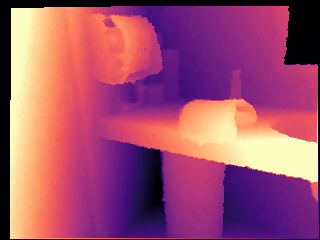}
    }\hspace{-8pt}
    \\ \vspace{-8pt}
    \subfloat{
        \includegraphics[width=.16\textwidth]{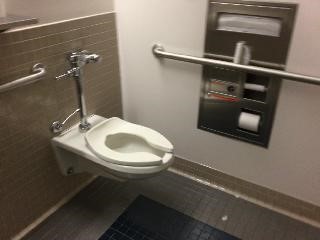}
    }\hspace{-9pt}
    \subfloat{
        \includegraphics[width=.16\textwidth]{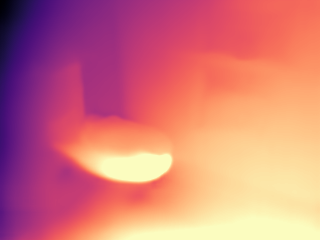}
    }\hspace{-8pt}
    \subfloat{
        \includegraphics[width=.16\textwidth]{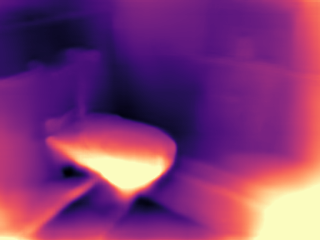}
    }\hspace{-8pt}
    \subfloat{
        \includegraphics[width=.16\textwidth]{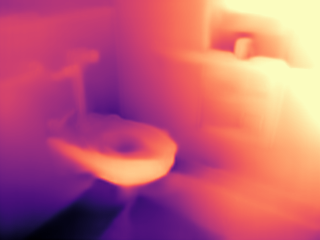}
    }\hspace{-8pt}
    \subfloat{
        \includegraphics[width=.16\textwidth]{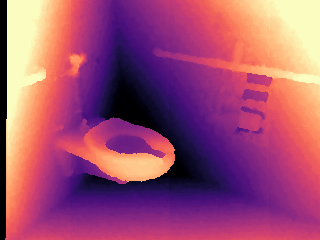}
    }\hspace{-8pt}
    \\ 
    \caption{Qualitative results on ScanNet dataset \cite{ScanNet}. The depth maps
    generated by our model have shaper edges and are closer to the
    ground-truth. Better viewed by zooming on screen.\label{fig:scannet}}
    % \vspace{-15pt}
\end{figure*}

Fig. \ref{fig:scannet} shows the visual results on ScanNet
\cite{ScanNet} dataset. The ground-truth depth maps in the last column are
interpolated from sparse point clouds for visualization purpose. We can
observe that our depth maps are more accurate and closer to the
ground-truth when compared with PLNet \cite{PLNet} and StructDepth \cite{StructDepth}. For
instance, on the area where there is a shelf in the first row, our depth
map can reflect its shape more clearly. In the third row, the
predictions of small objects in our depth maps are much shaper and
clearer. All these observations are consistent with the corresponding
quantitative results in Table \ref{tab:scannet}.
% \begin{table*}[!t!b]
%     \centering
    
% \end{table*}

\subsection{Ablation Study}

\begin{table*}[!htb]
\caption{Ablation Study. Quantitative Results for Different Variants of Our Model
(IndoorDepth) on NYUv2 Dataset \cite{NYUv2}.\label{tab:ablation}}
\centering
\renewcommand\arraystretch{1.2}
\begin{tabularx}{.85\textwidth}{lcc*{6}{C}}
% \hhline{|--||*{6}-|}
% \thickhline
\toprule
 &   &   & \multicolumn{3}{c}{Error Metric ↓} & \multicolumn{3}{c}{Accuracy Metric ↑}                                                   \\ \cmidrule(r){4-6} \cmidrule(r){7-9}   
\multirow{-2}{*}[2pt]{Ablation} & \multirow{-2}{*}[2pt]{Novel-Ph} & \multirow{-2}{*}[2pt]{Multi-Ph} & Abs Rel & Log10 & RMS & $\delta$ \textless 1.25 & $\delta$ \textless 1.25$^2$ & $\delta$ \textless 1.25$^3$ \\ %\hhline{|*{2}=::*{6}=|}
% \hhline{|-|-|-||*{6}{-|}}
\midrule
% \thickhline
PLNet \cite{PLNet} & & & 0.151 & 0.064 & 0.562 & 0.790 & 0.953 & 0.989 \\
Baseline (PLNet + Res1) & & & 0.132 & 0.057 & 0.516 & 0.831 & 0.961 &
0.990 \\
% \hhline{|-|-|-||*{6}{-|}}
\midrule
Baseline + Novel-Ph & \textbf{\checkmark} & & 0.129 & 0.055 & 0.506 & 0.838 &
0.964 & 0.991 \\
Baseline + Multi-Ph & & \textbf{\checkmark} & 0.128 & 0.055 & 0.503 & 0.840 &
0.963 & 0.990 \\
\textbf{IndoorDepth} & \textbf{\checkmark} & \textbf{\checkmark} & \textbf{0.126} &
\textbf{0.054} & \textbf{0.494} & \textbf{0.845} & \textbf{0.965} &
\textbf{0.991} \\
% \thickhline
\bottomrule
\end{tabularx}
\end{table*}

\begin{figure*}[h!tb]
    \centering
    \begin{tabularx}{.81\textwidth}{CCCC}
        Input\vspace{-12pt} & Baseline + Novel-Ph\vspace{-12pt} & Baseline + Multi-Ph\vspace{-12pt} & Ours (full model)\vspace{-12pt} \\
    \end{tabularx}
    % \vspace{-10pt}      
    \subfloat{
        \includegraphics[width=.2\textwidth]{6-1-1.png}
    }
    \hspace{-9pt}
    \subfloat{
        \includegraphics[width=.2\textwidth]{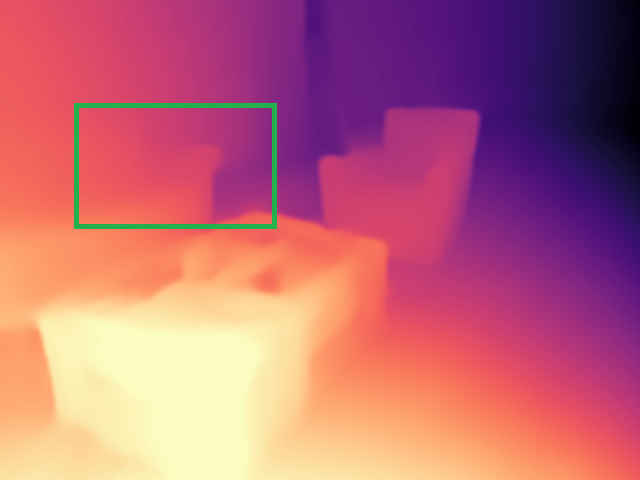}
    }
    \hspace{-8pt}
    \subfloat{
        \includegraphics[width=.2\textwidth]{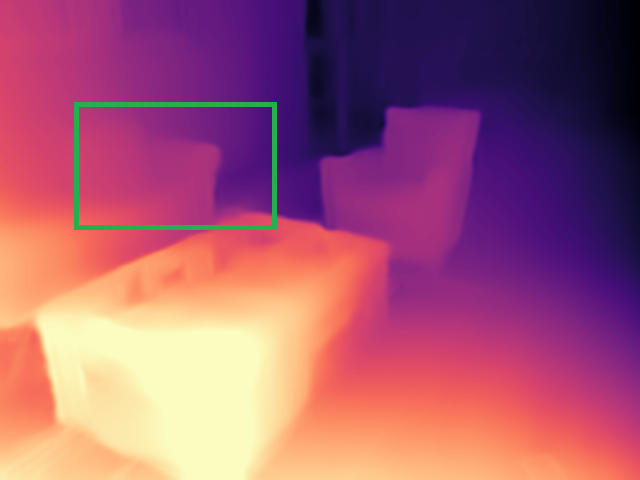}
    }\hspace{-8pt}
    \subfloat{
        \includegraphics[width=.2\textwidth]{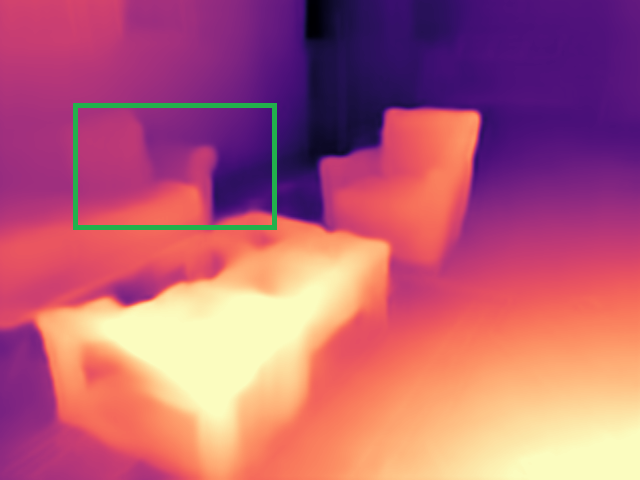}
    }
    \\ \vspace{-8pt}
    \subfloat{
        \includegraphics[width=.2\textwidth]{6-2-1.png}
    }\hspace{-9pt}
    \subfloat{
        \includegraphics[width=.2\textwidth]{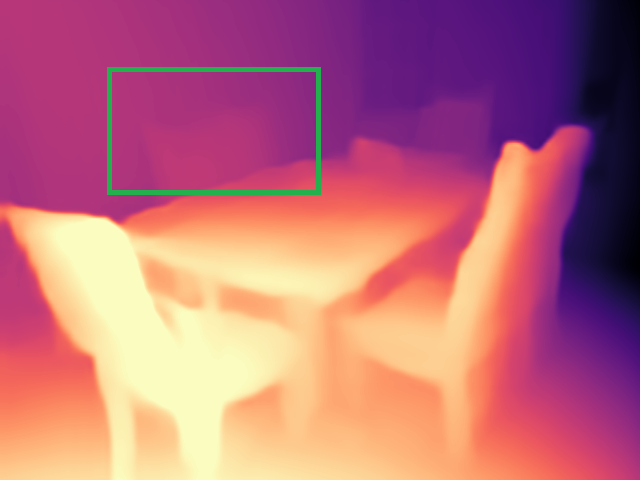}
    }\hspace{-8pt}
    \subfloat{
        \includegraphics[width=.2\textwidth]{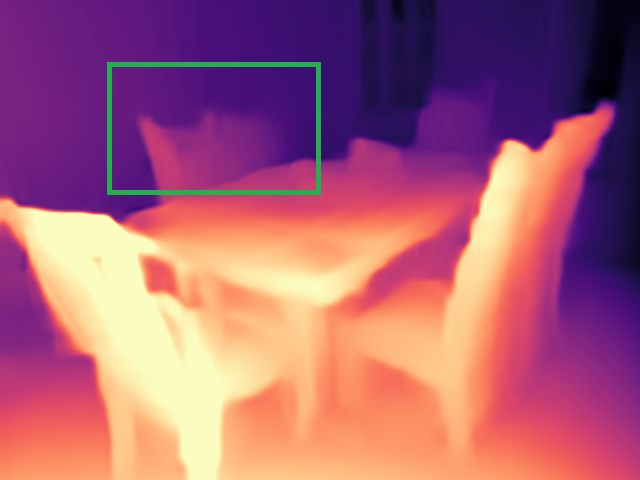}
    }\hspace{-8pt}
    \subfloat{
        \includegraphics[width=.2\textwidth]{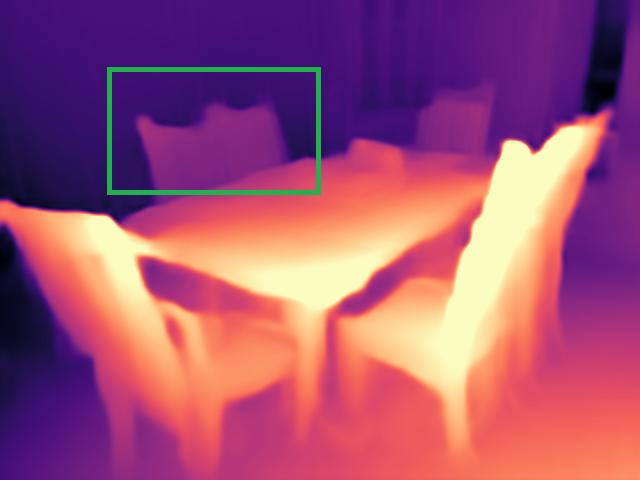}
    }
    \\ \vspace{-8pt}
    \subfloat{
        \includegraphics[width=.2\textwidth]{6-3-1.png}
    }\hspace{-9pt}
    \subfloat{
        \includegraphics[width=.2\textwidth]{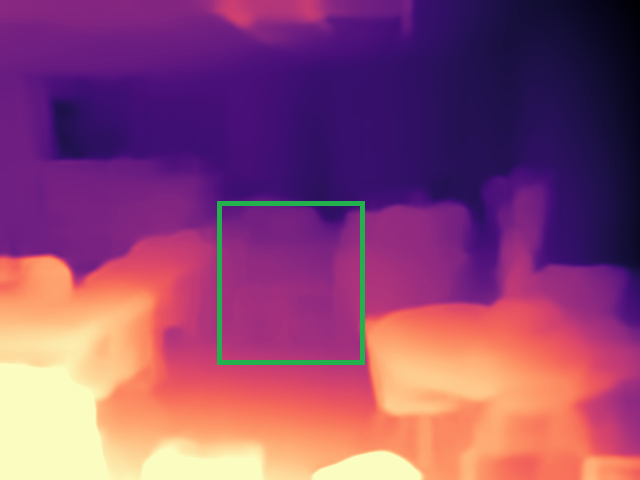}
    }\hspace{-8pt}
    \subfloat{
        \includegraphics[width=.2\textwidth]{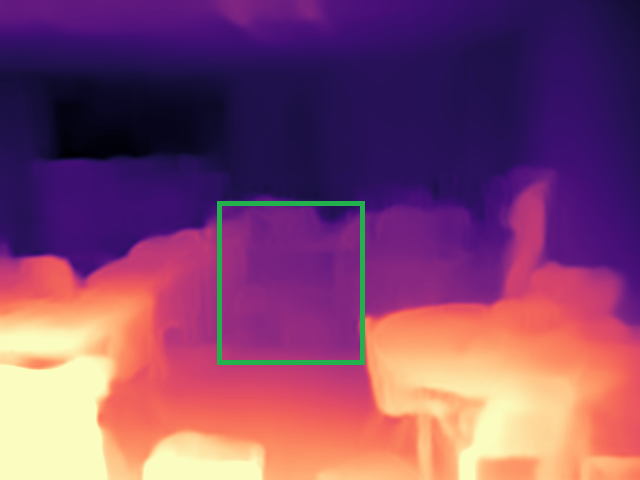}
    }\hspace{-8pt}
    \subfloat{
        \includegraphics[width=.2\textwidth]{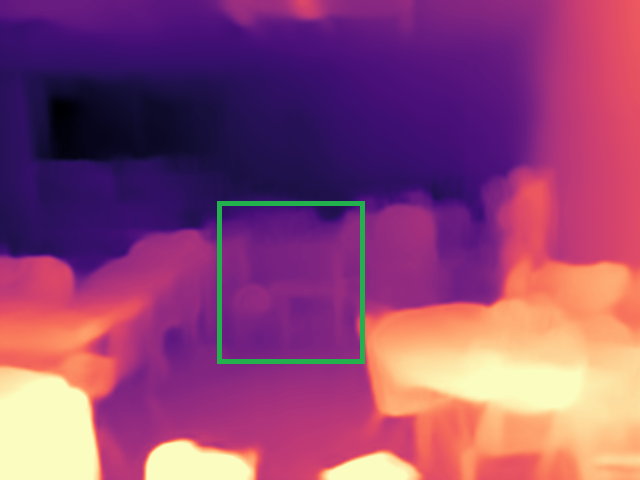}
    }
    \\ 
    \caption{Qualitative ablation results on NYUv2 dataset \cite{NYUv2}. Our full
    model produces better depth maps which have clearer object boundaries.
    Better viewed by zooming on screen.\label{fig:ablation}}
\end{figure*}

Table \ref{tab:ablation} presents the ablation results on NYUv2 dataset \cite{NYUv2}.
Novel-Ph denotes our novel photometric loss with improved \text{SSIM} in
Section \ref{sec:3B}. Multi-Ph means the deeper pose network trained by
multi-photometric losses in Section \ref{sec:3C}. Res1 in our baseline means
only one residual pose block. To compare more fairly, our baseline in
the second row is the combination of the depth network of PLNet \cite{PLNet}
and the pose network of MonoIndoor \cite{MonoIndoor}. We add our proposed Novel-Ph
module and Multi-Ph module in the third and fourth row, respectively.
Quantitative results demonstrate the effectiveness of each module when
compared with our baseline. Furthermore, the results in the last row
indicate that our full model, IndoorDepth, outperforms other models in
Table \ref{tab:ablation}. Then we compared three qualitative samples of ablation
results on NYUv2 dataset, as shown in Fig. \ref{fig:ablation}. The depth maps generated
by our full model have sharper edges than others. These observations can
prove the effectiveness of our modules, which align with the
quantitative results in Table \ref{tab:ablation}.

% \begin{table*}
%     \centering
%     % \vspace{-10pt}
    
% \end{table*}

\subsubsection{Effects of the proposed novel photometric loss (Novel-Ph)}

\begin{table}[!h!tb]
    \caption{Three Intuitive Approaches to Increasing the Value of \(L_{\text{ssim}}\).
    \(L_{\text{ssim1}}\) is the Structural Similarity Index of Method 1,
    \(L_{\text{ssim2}}\) is the Structural Similarity Index of Method 2 and
    \(L_{\text{ssim3}}\) is the Structural Similarity Index of Method 3.\label{tab:intuitive}}
    \centering
    \scriptsize
    % \footnotesize
    \renewcommand\arraystretch{2.5}
    % \resizebox{\linewidth}{10mm}{
    \begin{tabular}{|c|c|}
        \hline
        method 1 & \(L_{\text{ssim1}} = \frac{\alpha}{2}( 1 - (\text{SSIM}( I_{t},\ I_{t^{\prime} \rightarrow t} ) - \varepsilon) )\ \ (0 < \varepsilon < 1)\) \\ \hline
method 2 &
\(L_{\text{ssim2}} = \frac{\alpha}{2}( 1 - \rho \times \text{SSIM}( I_{t},\ I_{t^{\prime} \rightarrow t} ) )\ \ (0 < \rho < 1)\) \\ \hline
method 3 &
\(L_{\text{ssim3}} = \frac{\tau}{2}( 1 - \text{SSIM}( I_{t},\ I_{t^{\prime} \rightarrow t} ) )\ \ (\tau \geq \alpha)\) \\
\hline
    \end{tabular}
    % }
    % \vspace{-6ex}
\end{table}

In Section \ref{sec:3C}, we find that our proposed novel photometric loss with
improved \text{SSIM} can increase the value of \(L_{\text{ssim}}\) in Eq. (\ref{eq:8}) by
reducing the structural similarity index. To investigate whether the
performance of depth predictions can be improved by simply increasing
the value of \(L_{\text{ssim}}\), we also propose three intuitive approaches
for comparisons, as shown in Table \ref{tab:intuitive}. In method 1, we use
\(\text{SSIM}( I_{t},\ I_{t^{\prime} \rightarrow t} ) - \varepsilon\)
(\(0 < \varepsilon < 1\)) instead of
\(\text{SSIM}( I_{t},\ I_{t^{\prime} \rightarrow t} )\) to reduce the
value of structural similarity index and in turn increase \(L_{\text{ssim}}\).
Based on the equation of method 1, we have
% \vspace{-4ex}
\begin{align}
    L_{\text{ssim1}} &= \frac{\alpha}{2}( 1 - ( \text{SSIM}( I_{t},\ I_{t^{\prime} \rightarrow t} ) - \varepsilon ) )  \notag \\
% &= \frac{\alpha}{2}( 1 - \text{SSIM}( I_{t},\ I_{t^{\prime} \rightarrow t} ) + \varepsilon ) \notag \\
&= \frac{\alpha}{2}( 1 - \text{SSIM}( I_{t},\ I_{t^{\prime} \rightarrow t} ) ) + \frac{\alpha\varepsilon}{2} \notag \\
& = L_{\text{ssim}} + \frac{\alpha\varepsilon}{2} 
\end{align}
where \(\frac{\alpha\varepsilon}{2}\) is a constant. Thus it is clear
that method 1 does not have any effect on the training results. In
method 2, we increase the value of \(L_{\text{ssim}}\) by using constant
\(\rho\) (\(0 < \rho < 1\)) to decrease the value of structural
similarity index. Based on the equation of method 2, we have
\begin{align}
    L_{\text{ssim2}} &= \frac{\alpha}{2}( 1 - \rho \times \text{SSIM}( I_{t},\ I_{t^{\prime} \rightarrow t} ) ) \notag \\ 
    &= \frac{\alpha}{2}( 1 - \rho + \rho - \rho \times \text{SSIM}( I_{t},\ I_{t^{\prime} \rightarrow t} ) )\notag \\
    &= \frac{\alpha}{2}(1 - \rho) + \rho \times \frac{\alpha}{2}( 1 - \text{SSIM}( I_{t},\ I_{t^{\prime} \rightarrow t} ) ) \notag \\
    &= \frac{\alpha}{2}(1 - \rho) + {\rho L}_{\text{ssim}}
\end{align}

\begin{table*}[!tb]
    \caption{Effects of the Proposed Novel Photometric Loss (Novel-Ph).\label{tab:ablation1}}
    \centering
    \renewcommand\arraystretch{1.2}
    \begin{tabularx}{.85\textwidth}{l*{6}{C}}
   % \hhline{|--||*{6}-|}
    % \thickhline
    \toprule
           & \multicolumn{3}{c}{Error Metric ↓}                                                                              & \multicolumn{3}{c}{Accuracy Metric ↑}                                                   \\ \cmidrule(r){2-4} \cmidrule(r){5-7}
    \multirow{-2}{*}[2pt]{Methods}  & Abs Rel & Log10 & RMS & $\delta$ \textless 1.25 & $\delta$ \textless 1.25$^2$ & $\delta$ \textless 1.25$^3$ \\ %\hhline{|*{2}=::*{6}=|}
    % \hhline{|-||*{6}{-|}}
    \midrule
    % \thickhline
    Baseline1 (PLNet) \cite{PLNet} & 0.151 & 0.064 & 0.562 & 0.790 & 0.953 & 0.989 \\
    Baseline1 + Novel-Ph & 0.147 & 0.063 & 0.554 & 0.801 & 0.954 & 0.989 \\
% \hhline{|-||*{6}{-|}}
\midrule
Baseline2 (PLNet + Res1) & 0.132 & 0.057 & 0.516 & 0.831 & 0.961 & 0.990 \\
Baseline2 + Novel-Ph & \textbf{0.129} & \textbf{0.055} &
\textbf{0.506} & \textbf{0.838} & \textbf{0.964} & \textbf{0.991} \\
    % \thickhline
    \bottomrule
    \end{tabularx}
  \end{table*}

  \begin{table*}[!h!tb]
    \caption{Effects of the Proposed Deeper Pose Network (Multi-Ph). "No.
    Photometric Loss" Denotes the Number of Photometric Losses and "No.
    Residual Pose Block" Means the Number of Residual Pose Blocks in the
    Pose Network.\label{tab:ablation2}}
    \centering
    \renewcommand\arraystretch{1.2}
    \begin{tabularx}{.85\textwidth}{cc*{6}C}
    % \thickhline
    \toprule
         &           & \multicolumn{3}{c}{Error Metric ↓}                                                                              & \multicolumn{3}{c}{Accuracy Metric ↑}                                                   \\ \cmidrule(r){3-5} \cmidrule(r){6-8} 
         \multirow{-2}{*}[2pt]{\tabincell{c}{No. Photometric \\ Loss}} & \multirow{-2}{*}[2pt]{\tabincell{c}{No. Residual \\ Pose Block}} & Abs Rel & Log10 & RMS & $\delta$ \textless 1.25 & $\delta$ \textless 1.25$^2$ & $\delta$ \textless 1.25$^3$ \\ %\hhline{|*{2}=::*{6}=|}
    % \hhline{|-|-||*{6}{-|}}
    \midrule
    % \thickhline
    1 & 0 & 0.151 & 0.064 & 0.562 & 0.790 & 0.953 & 0.989 \\
    % \hhline{|-|-||*{6}{-|}}
    \midrule
1 & 1 & 0.132 & 0.057 & 0.516 & 0.831 & 0.961 & 0.990 \\
1 & 2 & 0.132 & 0.056 & 0.514 & 0.833 & 0.962 & \textbf{0.991} \\
1 & 3 & 0.133 & 0.057 & 0.519 & 0.827 & 0.960 & 0.990 \\
% \hhline{|-|-||*{6}{-|}}
\midrule
2 & 2 & \textbf{0.128} & \textbf{0.055} & \textbf{0.503} &
\textbf{0.840} & \textbf{0.963} & 0.990 \\
3 & 3 & 0.131 & 0.056 & 0.510 & 0.835 & 0.962 & 0.990 \\
    % \thickhline
    \bottomrule
    \end{tabularx}
\end{table*}

\begin{figure}[!h!tb]
    \centering
    \includegraphics[width=\linewidth]{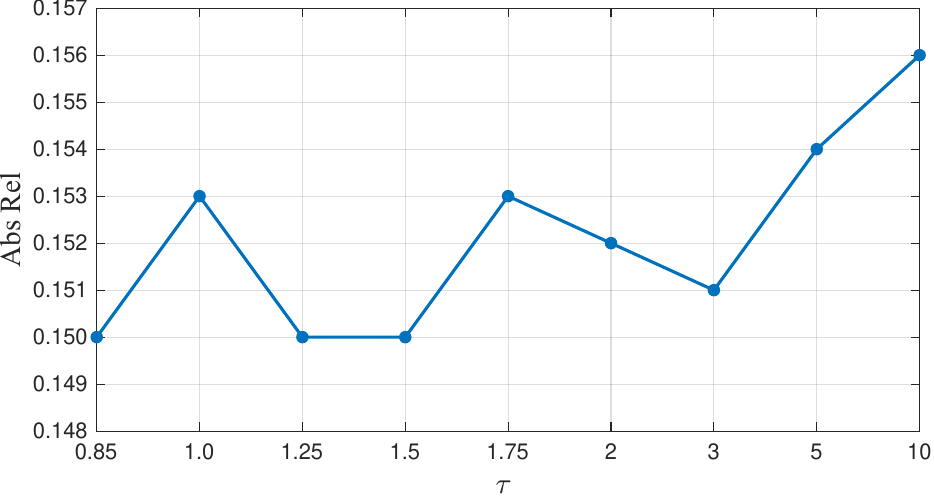}
    \caption{The change of AbsRel with $\tau$. Note that the weight of L1 loss $1-\alpha$ are fixed and the model is the same as PLNet when $\tau=\alpha=0.85$. However, we find that PLNet can reach an AbsRel of 0.150 after multiple runs. To compare more fairly, we use the AbsRel of 0.150 as the baseline here. \label{fig:method3}}
\end{figure}
It shows that \(L_{\text{ssim2}}\) consists of two parts, a constant term
\(\frac{\alpha}{2}(1 - \rho)\) that does not affect the training and a
reduced \(L_{\text{ssim}}\). Therefore, method 2 actually diminishes the impact
of structural similarity on training. In method 3, we adopt \(\tau\)
(\(\tau \geq \alpha\)) instead of \(\alpha\) to increase the value of
\(L_{\text{ssim}}\). However, the results of this approach in Fig. \ref{fig:method3} demonstrate that the performance does not improve or even becomes worse. In summary, all the intuitive approaches in Table \ref{tab:intuitive} fail to predict more precise depth. Therefore, it can be concluded that the performance of depth predictions cannot be improved by simply increasing the value of
\(L_{\text{ssim}}\).

In order to show the effects of Novel-Ph more richly, we use two
different baselines in Table \ref{tab:ablation1}. One is PLNet using a 3-frame input and
the other is the combination of PLNet and Res1. Specifically, compared
with PLNet in the first row, the model in the second row reduces AbsRel
by 0.4\%. Our model in the last row, which also adopts the proposed
Novel-Ph, reduces AbsRel by 0.3\% in comparison with the second
baseline. The results in Table \ref{tab:ablation1} indicate that our proposed novel
photometric loss can help us obtain more accurate depth predictions.

\subsubsection{Effects of the proposed deeper pose network (Multi-Ph)}

Table \ref{tab:ablation2} provides a more detailed comparison by varying the number of
photometric losses and residual pose blocks. The first row presents the
results of PLNet that does not have any residual pose blocks. Then we
simply add more residual pose blocks from the second to the fourth row.
It can be found from these three rows that the performance does not
significantly improve or even becomes worse when the number of residual
blocks is more than one. To handle the above issue, we adopt
multi-photometric losses to train the network in the last two rows.
Compared to the results in the second row, the performance is further
significantly improved after adopting the proposed deeper pose network
(Multi-Ph) in the fifth row. Furthermore, the results in the third and
fifth rows indicate that our proposed multi-photometric losses are
beneficial to the convergence of deeper pose network and further obtain
a better performance. The above observations can validate the
effectiveness of our proposed deeper pose network (Multi-Ph).

However, the pose networks with three residual pose blocks in the fourth
and last rows fail to improve the performance regardless of whether
multi-photometric losses are used. In fact, we report the best results
of multiple runs as the networks in the fourth and last rows sometimes
fail to converge. One possible reason is that the relative pose between
the input images of the last residual pose block is too small to provide
sufficient translation. The careful investigation of this phenomenon
will be leaved for our future work.

\section{Conclusion}
In this paper, we propose a self-supervised monocular indoor depth
estimation method, named IndoorDepth, including two novel strategies.
We first introduce our novel photometric loss with an improved \text{SSIM}
function which helps us to obtain more reasonable structural similarity
index for low-texture regions. Then a deeper pose network with two
residual pose blocks is designed to estimate more accurate camera poses.
In order to make all the added residual pose blocks work, we employ multiple photometric losses at different stages to train our network.
Extensive experiments demonstrate
that our model achieves state-of-the-art depth predictions among the
self-supervised methods.

In future work, we plan to design more advanced \text{SSIM} function to further
improve depth predictions in indoor environments. Moreover, since the
performance becomes worse when the number of residual pose blocks is
more than two, we will investigate the fundamental reasons and try to
solve this problem.

\bibliography{reff}

\begin{thebibliography}{10}
\providecommand{\url}[1]{#1}
\csname url@samestyle\endcsname
\providecommand{\newblock}{\relax}
\providecommand{\bibinfo}[2]{#2}
\providecommand{\BIBentrySTDinterwordspacing}{\spaceskip=0pt\relax}
\providecommand{\BIBentryALTinterwordstretchfactor}{4}
\providecommand{\BIBentryALTinterwordspacing}{\spaceskip=\fontdimen2\font plus
\BIBentryALTinterwordstretchfactor\fontdimen3\font minus
  \fontdimen4\font\relax}
\providecommand{\BIBforeignlanguage}[2]{{%
\expandafter\ifx\csname l@#1\endcsname\relax
\typeout{** WARNING: IEEEtran.bst: No hyphenation pattern has been}%
\typeout{** loaded for the language `#1'. Using the pattern for}%
\typeout{** the default language instead.}%
\else
\language=\csname l@#1\endcsname
\fi
#2}}
\providecommand{\BIBdecl}{\relax}
\BIBdecl

\bibitem{Autonomous1}
H.~Caesar, V.~Bankiti, A.~H. Lang, S.~Vora, V.~E. Liong, Q.~Xu, A.~Krishnan,
  Y.~Pan, G.~Baldan, and O.~Beijbom, ``nuscenes: A multimodal dataset for
  autonomous driving,'' in \emph{Proc. IEEE/CVF Conf. Comput. Vis. Pattern
  Recognit. (CVPR)}, Jun. 2020, pp. 11\,618--11\,628.

\bibitem{Autonomous2}
X.~Song, P.~Wang, D.~Zhou, R.~Zhu, C.~Guan, Y.~Dai, H.~Su, H.~Li, and R.~Yang,
  ``Apollocar3d: A large 3d car instance understanding benchmark for autonomous
  driving,'' in \emph{Proc. IEEE/CVF Conf. Comput. Vis. Pattern Recognit.
  (CVPR)}, Jun. 2019, pp. 5447--5457.

\bibitem{Autonomous3}
P.~Li, X.~Chen, and S.~Shen, ``Stereo r-cnn based 3d object detection for
  autonomous driving,'' in \emph{Proc. IEEE/CVF Conf. Comput. Vis. Pattern
  Recognit. (CVPR)}, Jun. 2019, pp. 7636--7644.

\bibitem{KITTI0}
A.~Geiger, P.~Lenz, and R.~Urtasun, ``Are we ready for autonomous driving? the
  kitti vision benchmark suite,'' in \emph{Proc. IEEE/CVF Conf. Comput. Vis.
  Pattern Recognit. (CVPR)}, Jun. 2012, pp. 3354--3361.

\bibitem{3D1}
M.~Tatarchenko, A.~Dosovitskiy, and T.~Brox, ``Octree generating networks:
  Efficient convolutional architectures for high-resolution 3d outputs,'' in
  \emph{Proc. IEEE/CVF Int. Conf. Comput. Vis. (ICCV)}, Oct. 2017, pp.
  2107--2115.

\bibitem{3D2}
J.~Mahmud, T.~Price, A.~Bapat, and J.-M. Frahm, ``Boundary-aware 3d building
  reconstruction from a single overhead image,'' in \emph{Proc. IEEE/CVF Conf.
  Comput. Vis. Pattern Recognit. (CVPR)}, Jun. 2020, pp. 438--448.

\bibitem{VR1}
E.~Bastug, M.~Bennis, M.~Medard, and M.~Debbah, ``Toward interconnected virtual
  reality: Opportunities, challenges, and enablers,'' \emph{IEEE Commun. Mag.},
  vol.~55, no.~6, pp. 110--117, Jun. 2017.

\bibitem{VR2}
F.~El~Jamiy and R.~Marsh, ``Survey on depth perception in head mounted
  displays: distance estimation in virtual reality, augmented reality, and
  mixed reality,'' \emph{IET Image Process.}, vol.~13, no.~5, pp. 707--712,
  Apr. 2019.

\bibitem{VR3}
M.-B. Ibáñez and C.~Delgado-Kloos, ``Augmented reality for stem learning: A
  systematic review,'' \emph{Comput. Educ.}, vol. 123, pp. 109--123, Aug. 2018.

\bibitem{Supervised1}
D.~Eigen and R.~Fergus, ``Predicting depth, surface normals and semantic labels
  with a common multi-scale convolutional architecture,'' in \emph{Proc.
  IEEE/CVF Int. Conf. Comput. Vis. (ICCV)}, Dec. 2015, pp. 2650--2658.

\bibitem{Supervised2}
A.~Chakrabarti, J.~Shao, and G.~Shakhnarovich, ``Depth from a single image by
  harmonizing overcomplete local network predictions,'' in \emph{Proc. Adv.
  Neural Inf. Process. Syst. (NIPS)}, D.~Lee, M.~Sugiyama, U.~Luxburg,
  I.~Guyon, and R.~Garnett, Eds., 2016, pp. 2666--2674.

\bibitem{Supervised3}
I.~Laina, C.~Rupprecht, V.~Belagiannis, F.~Tombari, and N.~Navab, ``Deeper
  depth prediction with fully convolutional residual networks,'' in \emph{Proc.
  Int. Conf. 3D Vis. (3DV)}, Oct. 2016, pp. 239--248.

\bibitem{Supervised4}
J.~Li, R.~Klein, and A.~Yao, ``A two-streamed network for estimating
  fine-scaled depth maps from single rgb images,'' in \emph{Proc. IEEE/CVF Int.
  Conf. Comput. Vis. (ICCV)}, Oct. 2017, pp. 3392--3400.

\bibitem{DORN}
H.~Fu, M.~Gong, C.~Wang, K.~Batmanghelich, and D.~Tao, ``Deep ordinal
  regression network for monocular depth estimation,'' in \emph{Proc. IEEE/CVF
  Conf. Comput. Vis. Pattern Recognit. (CVPR)}, Jun. 2018, pp. 2002--2011.

\bibitem{MD1}
C.~Godard, O.~Mac~Aodha, and G.~J. Brostow, ``Unsupervised monocular depth
  estimation with left-right consistency,'' in \emph{Proc. IEEE/CVF Conf.
  Comput. Vis. Pattern Recognit. (CVPR)}, Jul. 2017, pp. 6602--6611.

\bibitem{MD2}
C.~Godard, O.~Mac~Aodha, M.~Firman, and G.~Brostow, ``Digging into
  self-supervised monocular depth estimation,'' in \emph{Proc. IEEE/CVF Int.
  Conf. Comput. Vis. (ICCV)}, Oct. 2019, pp. 3827--3837.

\bibitem{Semantic1}
V.~Guizilini, R.~Hou, J.~Li, R.~Ambrus, and A.~Gaidon, ``Semantically-guided
  representation learning for self-supervised monocular depth,'' in \emph{Proc.
  ICLR}, 2020.

\bibitem{DDV}
A.~Johnston and G.~Carneiro, ``Self-supervised monocular trained depth
  estimation using self-attention and discrete disparity volume,'' in
  \emph{Proc. IEEE/CVF Conf. Comput. Vis. Pattern Recognit. (CVPR)}, Jun. 2020,
  pp. 4755--4764.

\bibitem{PackNet}
V.~Guizilini, R.~Ambrus, S.~Pillai, A.~Raventos, and A.~Gaidon, ``3d packing
  for self-supervised monocular depth estimation,'' in \emph{Proc. IEEE/CVF
  Conf. Comput. Vis. Pattern Recognit. (CVPR)}, Jun. 2020, pp. 2482--2491.

\bibitem{Fine-grained}
H.~Jung, E.~Park, and S.~Yoo, ``Fine-grained semantics-aware representation
  enhancement for self-supervised monocular depth estimation,'' in \emph{Proc.
  IEEE/CVF Int. Conf. Comput. Vis. (ICCV)}, Oct. 2021, pp. 12\,622--12\,632.

\bibitem{KITTI}
A.~Geiger, P.~Lenz, C.~Stiller, and R.~Urtasun, ``Vision meets robotics: The
  kitti dataset,'' \emph{Int. J. Robot. Res.}, vol.~32, no.~11, pp. 1231--1237,
  Sep. 2013.

\bibitem{Make3D}
A.~Saxena, M.~Sun, and A.~Y. Ng, ``Make3d: Learning 3d scene structure from a
  single still image,'' \emph{IEEE Trans. Pattern Anal. Mach. Intell.},
  vol.~31, no.~5, pp. 824--840, May 2009.

\bibitem{NYUv2}
N.~Silberman, D.~Hoiem, P.~Kohli, and R.~Fergus, ``Indoor segmentation and
  support inference from rgbd images,'' in \emph{Proc. Eur. Conf. Comput. Vis.
  (ECCV)}, Oct. 2012, pp. 746--760.

\bibitem{ScanNet}
A.~Dai, A.~X. Chang, M.~Savva, M.~Halber, T.~Funkhouser, and M.~Niessner,
  ``Scannet: Richly-annotated 3d reconstructions of indoor scenes,'' in
  \emph{Proc. IEEE/CVF Conf. Comput. Vis. Pattern Recognit. (CVPR)}, Jul. 2017,
  pp. 2432--2443.

\bibitem{feature-metric}
C.~Shu, K.~Yu, Z.~Duan, and K.~Yang, ``Feature-metric loss for self-supervised
  learning of depth and egomotion,'' in \emph{Proc. Eur. Conf. Comput. Vis.
  (ECCV)}, Aug. 2020, pp. 572--588.

\bibitem{ARN}
J.-W. Bian, H.~Zhan, N.~Wang, T.-J. Chin, C.~Shen, and I.~Reid, ``Auto-rectify
  network for unsupervised indoor depth estimation,'' \emph{IEEE Trans. Pattern
  Anal. Mach. Intell.}, pp. 1--1, 2021.

\bibitem{SSIM}
Z.~Wang, A.~Bovik, H.~Sheikh, and E.~Simoncelli, ``Image quality assessment:
  From error visibility to structural similarity,'' \emph{IEEE Trans. Image
  Process.}, vol.~13, no.~4, pp. 600--612, Apr. 2004.

\bibitem{MonoIndoor}
P.~Ji, R.~Li, B.~Bhanu, and Y.~Xu, ``Monoindoor: Towards good practice of
  self-supervised monocular depth estimation for indoor environments,'' in
  \emph{Proc. IEEE/CVF Int. Conf. Comput. Vis. (ICCV)}, Oct. 2021, pp.
  12\,767--12\,776.

\bibitem{Eigen1}
D.~Eigen, C.~Puhrsch, and R.~Fergus, ``Depth map prediction from a single image
  using a multi-scale deep network,'' in \emph{Proc. Adv. Neural Inf. Process.
  Syst. (NIPS)}, Dec. 2014.

\bibitem{Supervised6}
F.~Liu, C.~Shen, G.~Lin, and I.~Reid, ``Learning depth from single monocular
  images using deep convolutional neural fields,'' \emph{IEEE Trans. Pattern
  Anal. Mach. Intell.}, vol.~38, no.~10, pp. 2024--2039, Oct. 2016.

\bibitem{DFDepth}
Y.~Cao, Z.~Wu, and C.~Shen, ``Estimating depth from monocular images as
  classification using deep fully convolutional residual networks,'' \emph{IEEE
  Trans. Circuits Syst. Video Technol.}, vol.~28, no.~11, pp. 3174--3182, Nov.
  2018.

\bibitem{Pyramid-Depth}
M.~Song, S.~Lim, and W.~Kim, ``Monocular depth estimation using laplacian
  pyramid-based depth residuals,'' \emph{IEEE Trans. Circuits Syst. Video
  Technol.}, vol.~31, no.~11, pp. 4381--4393, Oct. 2021.

\bibitem{CORNet}
X.~Meng, C.~Fan, Y.~Ming, and H.~Yu, ``Cornet: Context-based ordinal regression
  network for monocular depth estimation,'' \emph{IEEE Trans. Circuits Syst.
  Video Technol.}, vol.~32, no.~7, pp. 4841--4853, Jul. 2022.

\bibitem{Supervised9}
J.~Hu, M.~Ozay, Y.~Zhang, and T.~Okatani, ``Revisiting single image depth
  estimation: Toward higher resolution maps with accurate object boundaries,''
  in \emph{IEEE Winter Conf. Appl. Comput. Vis. (WACV)}, Jan. 2019, pp.
  1043--1051.

\bibitem{Supervised10}
W.~Yin, Y.~Liu, C.~Shen, and Y.~Yan, ``Enforcing geometric constraints of
  virtual normal for depth prediction,'' in \emph{Proc. IEEE/CVF Int. Conf.
  Comput. Vis. (ICCV)}, Oct. 2019, pp. 5683--5692.

\bibitem{ZhouT}
T.~Zhou, M.~Brown, N.~Snavely, and D.~G. Lowe, ``Unsupervised learning of depth
  and ego-motion from video,'' in \emph{IEEE Conf. Comput. Vis. Pattern
  Recognit. (CVPR)}, Jul. 2017, pp. 6612--6619.

\bibitem{motionnet}
\BIBentryALTinterwordspacing
H.~Li, A.~Gordon, H.~Zhao, V.~Casser, and A.~Angelova, ``Unsupervised monocular
  depth learning in dynamic scenes,'' 2020, arXiv:2010.16404. [Online].
  Available: \url{https://arxiv.org/abs/2010.16404}
\BIBentrySTDinterwordspacing

\bibitem{HRDepth}
X.~Lyu, L.~Liu, M.~Wang, X.~Kong, L.~Liu, Y.~Liu, X.~Chen, and Y.~Yuan,
  ``Hr-depth: High resolution self-supervised monocular depth estimation,'' in
  \emph{Proc Conf AAAI Artif Intell.}, Feb. 2021, pp. 2294--2301.

\bibitem{FD-Photometric}
S.~Chen, Z.~Pu, X.~Fan, and B.~Zou, ``Fixing defect of photometric loss for
  self-supervised monocular depth estimation,'' \emph{IEEE Trans. Circuits
  Syst. Video Technol.}, vol.~32, no.~3, pp. 1328--1338, Mar. 2022.

\bibitem{DontForget}
V.~Patil, W.~Van~Gansbeke, D.~Dai, and L.~Van~Gool, ``Don’t forget the past:
  Recurrent depth estimation from monocular video,'' \emph{IEEE Robot. Autom.
  Lett.}, vol.~5, no.~4, pp. 6813--6820, Oct. 2020.

\bibitem{ManyDepth}
J.~Watson, O.~Mac~Aodha, V.~Prisacariu, G.~Brostow, and M.~Firman, ``The
  temporal opportunist: Self-supervised multi-frame monocular depth,'' in
  \emph{IEEE/CVF Conf. Comput. Vis. Pattern Recognit. (CVPR)}, Jun. 2021, pp.
  1164--1174.

\bibitem{MovingIndoor}
J.~Zhou, Y.~Wang, K.~Qin, and W.~Zeng, ``Moving indoor: Unsupervised video
  depth learning in challenging environments,'' in \emph{Proc. IEEE/CVF Int.
  Conf. Comput. Vis. (ICCV)}, Oct. 2019, pp. 8617--8626.

\bibitem{P2Net}
Z.~Yu, L.~Jin, and S.~Gao, ``P\textsuperscript{2}net: Patch-match and
  plane-regularization for unsupervised indoor depth estimation,'' in
  \emph{Proc. Eur. Conf. Comput. Vis. (ECCV)}, Aug. 2020, pp. 206--222.

\bibitem{StructDepth}
B.~Li, Y.~Huang, Z.~Liu, D.~Zou, and W.~Yu, ``Structdepth: Leveraging the
  structural regularities for self-supervised indoor depth estimation,'' in
  \emph{Proc. IEEE/CVF Int. Conf. Comput. Vis. (ICCV)}, Oct. 2021, pp.
  12\,663--12\,673.

\bibitem{PLNet}
H.~Jiang, L.~Ding, J.~Hu, and R.~Huang, ``Plnet: Plane and line priors for
  unsupervised indoor depth estimation,'' in \emph{Proc. Int. Conf. 3D Vis.
  (3DV)}, Dec. 2021, pp. 741--750.

\bibitem{MonoIndoor++}
R.~Li, P.~Ji, Y.~Xu, and B.~Bhanu, ``Monoindoor++:towards better practice of
  self-supervised monocular depth estimation for indoor environments,''
  \emph{IEEE Trans. Circuits Syst. Video Technol.}, pp. 1--1, 2022.

\bibitem{SSIML1}
H.~Zhao, O.~Gallo, I.~Frosio, and J.~Kautz, ``Loss functions for image
  restoration with neural networks,'' \emph{IEEE Trans. Comput. Imag}, vol.~3,
  no.~1, pp. 47--57, Mar. 2017.

\bibitem{Adam}
\BIBentryALTinterwordspacing
D.~P. Kingma and J.~Ba, ``Adam: A method for stochastic optimization,'' 2014,
  arXiv:1412.6980. [Online]. Available: \url{http://arxiv.org/abs/1412.6980}
\BIBentrySTDinterwordspacing

\bibitem{DepthTransfer}
K.~Karsch, C.~Liu, and S.~B. Kang, ``Depthtransfer: Depth extraction from video
  using non-parametric sampling,'' \emph{IEEE Trans. Pattern Anal. Mach.
  Intell.}, vol.~36, no.~11, pp. 2144--2158, Nov. 2014.

\bibitem{Li}
B.~Li, C.~Shen, Y.~Dai, A.~van~den Hengel, and M.~He, ``Depth and surface
  normal estimation from monocular images using regression on deep features and
  hierarchical crfs,'' in \emph{IEEE Conf. Comput. Vis. Pattern Recognit.
  (CVPR)}, Jun. 2015, pp. 1119--1127.

\bibitem{Wang}
P.~Wang, X.~Shen, Z.~Lin, S.~Cohen, B.~Price, and A.~Yuille, ``Towards unified
  depth and semantic prediction from a single image,'' in \emph{IEEE Conf.
  Comput. Vis. Pattern Recognit. (CVPR)}, Jun. 2015, pp. 2800--2809.

\bibitem{Supervised7}
D.~Xu, E.~Ricci, W.~Ouyang, X.~Wang, and N.~Sebe, ``Multi-scale continuous crfs
  as sequential deep networks for monocular depth estimation,'' in \emph{IEEE
  Conf. Comput. Vis. Pattern Recognit. (CVPR)}, Jul. 2017, pp. 161--169.

\bibitem{Fang}
Z.~Fang, X.~Chen, Y.~Chen, and L.~Van~Gool, ``Towards good practice for
  cnn-based monocular depth estimation,'' in \emph{IEEE Winter Conf. Appl.
  Comput. Vis. (WACV)}, Mar. 2020, pp. 1080--1089.

\bibitem{DistDepth}
C.~Y. Wu, J.~Wang, M.~Hall, U.~Neumann, and S.~Su, ``Toward practical monocular
  indoor depth estimation,'' in \emph{IEEE Conf. Comput. Vis. Pattern Recognit.
  (CVPR)}, Jun. 2022, pp. 3814--3824.

\bibitem{TrainFlow}
W.~Zhao, S.~Liu, Y.~Shu, and Y.-J. Liu, ``Towards better generalization: Joint
  depth-pose learning without posenet,'' in \emph{IEEE/CVF Conf. Comput. Vis.
  Pattern Recognit. (CVPR)}, Jun. 2020, pp. 9148--9158.

\bibitem{IFMNet}
Y.~Wei, H.~Guo, J.~Lu, and J.~Zhou, ``Iterative feature matching for
  self-supervised indoor depth estimation,'' \emph{IEEE Trans. Circuits Syst.
  Video Technol.}, vol.~32, no.~6, pp. 3839--3852, Jun. 2022.

\bibitem{DeepV2D}
\BIBentryALTinterwordspacing
Z.~Teed and J.~Deng, ``Deepv2d: Video to depth with differentiable structure
  from motion,'' 2018, arXiv:1812.04605. [Online]. Available:
  \url{http://arxiv.org/abs/1812.04605}
\BIBentrySTDinterwordspacing

\bibitem{SC-Depth}
J.-W. Bian, H.~Zhan, N.~Wang, Z.~Li, L.~Zhang, C.~Shen, M.-M. Cheng, and
  I.~Reid, ``Unsupervised scale-consistent depth learning from video,''
  \emph{Int. J. Comput. Vis.}, vol. 129, no.~9, pp. 2548--2564, Sep. 2021.

\end{thebibliography}
\bibliographystyle{IEEEtran}

\vfill

\end{document}

\title{Supplementary Material}

\maketitle

\vspace{-7ex}
\begin{table}[!h]
    \centering
    \normalsize
    % \emph{\url{https://github.com/fcntes/IndoorDepth}}
    \emph{\href{https://github.com/fcntes/IndoorDepth}{https://github.com/fcntes/IndoorDepth}}
\end{table}
\subsection{Proofs of Eq. (13), Eq. (14) and Eq. (15) in Section III-B}
\vspace{2ex}
The following is a simple proof of Eq. (13). For \(\text{SSIM}\left(I_{kx},\ I_{ky}\right)\), we have
\begin{align}
    \text{SSIM}\left( I_{kx},\ I_{ky} \right) &= l\left( I_{kx},\ I_{ky} \right)cs\left( I_{kx},\ I_{ky} \right) \notag \\
&= \frac{2\mu_{kx}\mu_{ky} + {(M_{1} \cdot kL)}^{2}}{\mu_{kx}^{2} + \mu_{ky}^{2} + {(M_{1} \cdot kL)}^{2}} \cdot \frac{2\sigma_{kxky} + {(M_{2} \cdot kL)}^{2}}{\sigma_{kx}^{2} + \sigma_{ky}^{2} + {(M_{2} \cdot kL)}^{2}} \notag \\
&= \frac{2{k\mu}_{x}{k\mu}_{y} + {(M_{1} \cdot kL)}^{2}}{{({k\mu}_{x})}^{2} + {({k\mu}_{y})}^{2} + {(M_{1} \cdot kL)}^{2}} \cdot \frac{2k^{2}\sigma_{xy} + {(M_{2} \cdot kL)}^{2}}{k^{2}\sigma_{x}^{2} + {k^{2}\sigma}_{y}^{2} + {(M_{2} \cdot kL)}^{2}} \notag \\
&= \frac{k^{2}(2\mu_{x}\mu_{y} + \left( M_{1} \cdot L \right)^{2})}{k^{2}(\mu_{x}^{2} + \mu_{y}^{2} + \left( M_{1} \cdot L \right)^{2})} \cdot \frac{k^{2}(2\sigma_{xy} + \left( M_{2} \cdot L \right)^{2})}{k^{2}(\sigma_{x}^{2} + \sigma_{y}^{2} + \left( M_{2} \cdot L \right)^{2})}\notag \\
&= l\left( I_{x},\ I_{y} \right)cs\left( I_{x},\ I_{y} \right) = \text{SSIM}\left( I_{x},\ I_{y} \right). \tag{27}
\end{align}
Then for the proof of Eq. (14), since
\(\text{SSIM}( I_{x},\ I_{y} ) = \text{SSIM}( I_{kx},\ I_{ky})\)
and \(k > 1\), we have
\begin{align}
    \text{SSIM}\left( I_{x},\ I_{y} \right) &= \text{SSIM}\left( I_{kx},\ I_{ky} \right)\notag \\
&= l\left( I_{kx},\ I_{ky} \right)cs\left( I_{kx},\ I_{ky} \right)\notag \\
&= \frac{2\mu_{kx}\mu_{ky} + {(M_{1} \cdot kL)}^{2}}{\mu_{kx}^{2} + \mu_{ky}^{2} + {(M_{1} \cdot kL)}^{2}} \cdot \frac{2\sigma_{kxky} + {(M_{2} \cdot kL)}^{2}}{\sigma_{kx}^{2} + \sigma_{ky}^{2} + {(M_{2} \cdot kL)}^{2}}\notag \\
&= \frac{2\mu_{kx}\mu_{ky} + {(M_{1} \cdot L)}^{2} + (k^{2} - 1){(M_{1} \cdot L)}^{2}}{\mu_{kx}^{2} + \mu_{ky}^{2} + {(M_{1} \cdot L)}^{2} + (k^{2} - 1){(M_{1} \cdot L)}^{2}} \cdot \frac{2\sigma_{kxky} + {(M_{2} \cdot L)}^{2} + (k^{2} - 1){(M_{2} \cdot L)}^{2}}{\sigma_{kx}^{2} + \sigma_{ky}^{2} + {(M_{2} \cdot L)}^{2} + (k^{2} - 1){(M_{2} \cdot L)}^{2}}, \tag{28}
\end{align}
and since \(\mu_{kx}^{2} + \mu_{ky}^{2} \geq 2\mu_{kx}\mu_{ky}\),
\(\sigma_{kx}^{2} + \sigma_{ky}^{2} \geq 2\sigma_{kxky}\) and \(k > 1\),
we have
\begin{align}
    \text{SSIM}^{\prime}\left( I_{x},\ I_{y} \right) &= l^{\prime}\left( I_{x},\ I_{y} \right){cs}^{\prime}\left( I_{x},\ I_{y} \right)\notag \\
&= \frac{2\mu_{kx}\mu_{ky} + \left( M_{1} \cdot L \right)^{2}}{\mu_{kx}^{2} + \mu_{ky}^{2} + \left( M_{1} \cdot L \right)^{2}} \cdot \frac{2\sigma_{kxky} + \left( M_{2} \cdot L \right)^{2}}{\sigma_{kx}^{2} + \sigma_{ky}^{2} + \left( M_{2} \cdot L \right)^{2}}\notag \\
&\leq \frac{2\mu_{kx}\mu_{ky} + {(M_{1} \cdot L)}^{2} + (k^{2} - 1){(M_{1} \cdot L)}^{2}}{\mu_{kx}^{2} + \mu_{ky}^{2} + {(M_{1} \cdot L)}^{2} + (k^{2} - 1){(M_{1} \cdot L)}^{2}} \cdot \frac{2\sigma_{kxky} + {(M_{2} \cdot L)}^{2} + (k^{2} - 1){(M_{2} \cdot L)}^{2}}{\sigma_{kx}^{2} + \sigma_{ky}^{2} + {(M_{2} \cdot L)}^{2} + (k^{2} - 1){(M_{2} \cdot L)}^{2}}. \tag{29}
\end{align}
Therefore, from Eq. (28) and Eq. (29), we have
\begin{equation}
    {\text{SSIM}}^{\prime} \left( I_{x},\ I_{y} \right) \leq \text{SSIM}\left( I_{x},\ I_{y} \right), \tag{30}
\end{equation}
which is the same as Eq. (14). If and only if \(I_{x} = I_{y}\), we have
\({\text{SSIM}}^{\prime} \left( I_{x},\ I_{y} \right) = \text{SSIM}\left( I_{x},\ I_{y} \right)\).
The following is the proof of Eq. (15).
For fixed input images
\(I_{x}\) and \(I_{y}\) and the variable \(k\) (\(k > 1\)), we have
\begin{align}
    {\text{SSIM}}_{k}^{\prime} \left( I_{x},\ I_{y} \right) &= \frac{2\mu_{kx}\mu_{ky} + \left( M_{1} \cdot L \right)^{2}}{\mu_{kx}^{2} + \mu_{ky}^{2} + \left( M_{1} \cdot L \right)^{2}} \cdot \frac{2\sigma_{kxky} + \left( M_{2} \cdot L \right)^{2}}{\sigma_{kx}^{2} + \sigma_{ky}^{2} + \left( M_{2} \cdot L \right)^{2}}\notag \\
&= \frac{2{k\mu}_{x}{k\mu}_{y} + {(M_{1}L)}^{2}}{{({k\mu}_{x})}^{2} + {({k\mu}_{y})}^{2} + {(M_{1}L)}^{2}} \cdot \frac{2k^{2}\sigma_{xy} + {(M_{2}L)}^{2}}{k^{2}\sigma_{x}^{2} + {k^{2}\sigma}_{y}^{2} + {(M_{2}L)}^{2}}\notag \\
% &= \frac{2{k\mu}_{x}{k\mu}_{y} + {(M_{1}L)}^{2} + {({k\mu}_{x})}^{2} + {({k\mu}_{y})}^{2} - \left( {k\mu}_{x} \right)^{2} - {({k\mu}_{y})}^{2}}{{({k\mu}_{x})}^{2} + {({k\mu}_{y})}^{2} + {(M_{1}L)}^{2}} \cdot \frac{2k^{2}\sigma_{xy} + {(M_{2}L)}^{2} + k^{2}\sigma_{x}^{2} + {k^{2}\sigma}_{y}^{2} - k^{2}\sigma_{x}^{2} - {k^{2}\sigma}_{y}^{2}}{k^{2}\sigma_{x}^{2} + {k^{2}\sigma}_{y}^{2} + {(M_{2}L)}^{2}}\notag \\
&= \left(1 - \frac{k^{2}(\mu_{x}^{2} + \mu_{y}^{2} - 2\mu_{x}\mu_{y})}{\left( {k\mu}_{x} \right)^{2} + {({k\mu}_{y})}^{2} + \left( M_{1}L \right)^{2}}\right) \cdot \left(1 - \frac{k^{2}(\sigma_{x}^{2} + \sigma_{y}^{2} - 2\sigma_{xy})}{k^{2}\sigma_{x}^{2} + {k^{2}\sigma}_{y}^{2} + \left( M_{2}L \right)^{2}}\right)\notag \\
&= \left(1 - \frac{\mu_{x}^{2} + \mu_{y}^{2} - 2\mu_{x}\mu_{y}}{\mu_{x}^{2} + \mu_{y}^{2} + \left( M_{1}L/k \right)^{2}}\right) \cdot \left(1 - \frac{\sigma_{x}^{2} + \sigma_{y}^{2} - 2\sigma_{xy}}{\sigma_{x}^{2} + \sigma_{y}^{2} + \left( M_{2}L/k \right)^{2}}\right)\tag{31}
\end{align}
and since \(\mu_{x}^{2} + \mu_{y}^{2} \geq 2\mu_{x}\mu_{y}\) and
\(\sigma_{x}^{2} + \sigma_{y}^{2} \geq 2\sigma_{xy}\), we can easily
conclude from Eq. (31) that a lower \(k\) can get a higher
\({\text{SSIM}}_{k}^{\prime} \left( I_{x},\ I_{y} \right)\). Therefore, for
\(k_{1} > k_{2}\), we have
\begin{equation}
    {\text{SSIM}}_{k_{1}}^{\prime} \left( I_{x},\ I_{y} \right) \leq {\text{SSIM}}_{k_{2}}^{\prime} \left( I_{x},\ I_{y} \right). \tag{32}
\end{equation}
If and only if \(I_{x} = I_{y}\), we have
\({\text{SSIM}}_{k_{1}}^{\prime} \left( I_{x},\ I_{y} \right) = {\text{SSIM}}_{k_{2}}^{\prime} \left( I_{x},\ I_{y} \right)\).

\newpage
\subsection{Point Cloud Visualization}
% \vspace{2ex}

\captionsetup[figure]{name={Fig.},labelsep=period,singlelinecheck=off}
\renewcommand{\thefigure}{8} 
\begin{figure}[!t]
    % \centering
    \hspace{.06\textwidth} Input(NYUv2)\vspace{-1pt} \hspace{.16\textwidth} 3D \vspace{-1pt} 
    \hfill
     Input(ScanNet)\vspace{-1pt} \hspace{.15\textwidth} 3D\vspace{-1pt} \hspace{.11\textwidth} \\
     \centering
    \subfloat{
        \includegraphics[width=.23\textwidth]{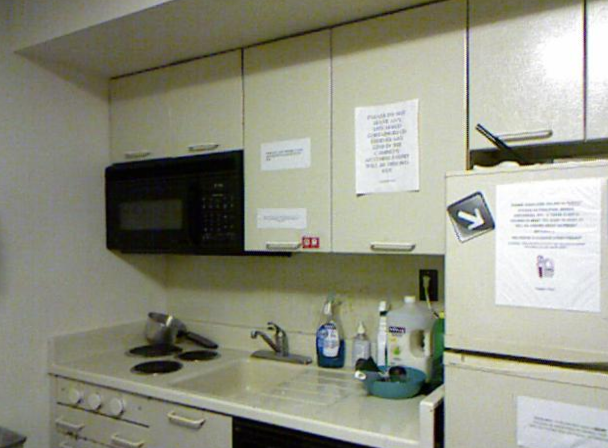}
    }
    \hspace{-5pt}
    \subfloat{
        \includegraphics[width=.23\textwidth]{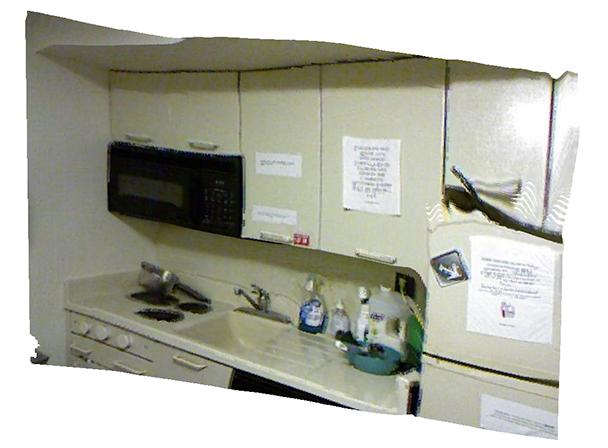}
    }
    \hfill
    \subfloat{
        \includegraphics[width=.23\textwidth]{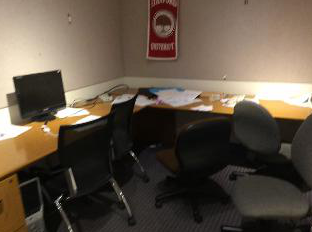}
    }\hspace{-5pt}
    \subfloat{
        \includegraphics[width=.23\textwidth]{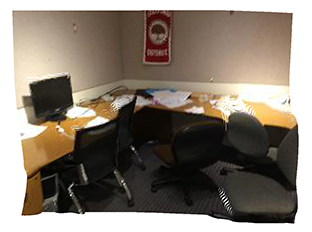}
    }
    \\ \vspace{-5pt}
    \subfloat{
        \includegraphics[width=.23\textwidth]{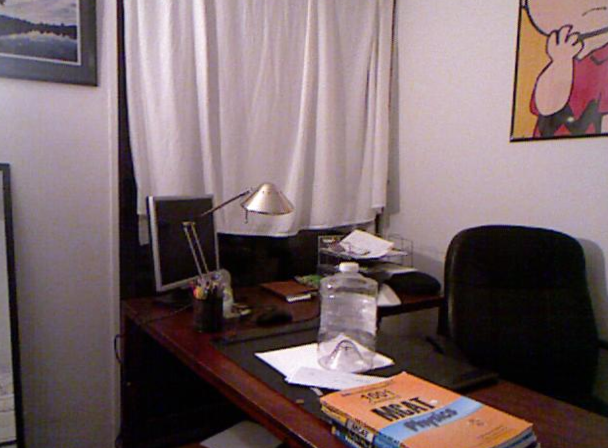}
    }\hspace{-5pt}
    \subfloat{
        \includegraphics[width=.23\textwidth]{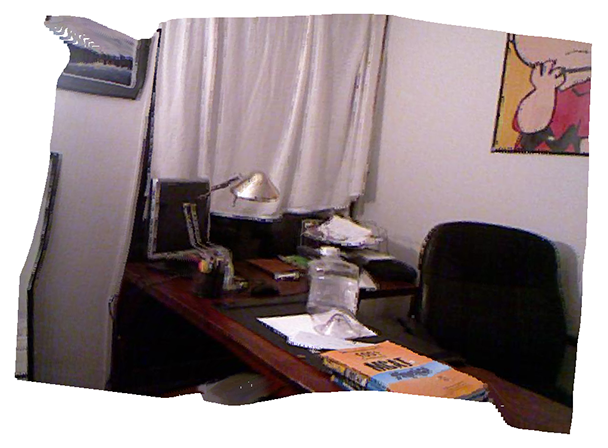}
    }
    \hfill
    \subfloat{
        \includegraphics[width=.23\textwidth]{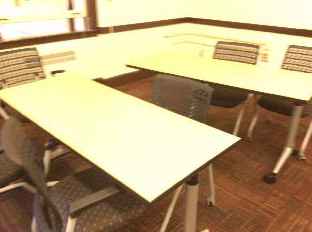}
    }\hspace{-5pt}
    \subfloat{
        \includegraphics[width=.23\textwidth]{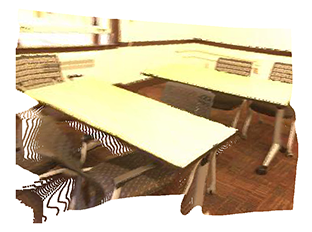}
    }
    \\ \vspace{-5pt}
    \subfloat{
        \includegraphics[width=.23\textwidth]{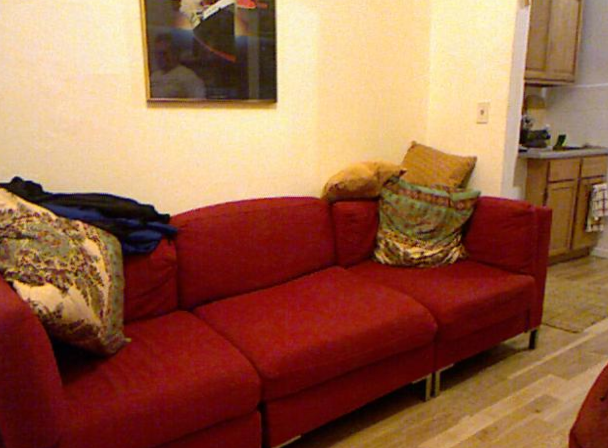}
    }\hspace{-5pt}
    \subfloat{
        \includegraphics[width=.23\textwidth]{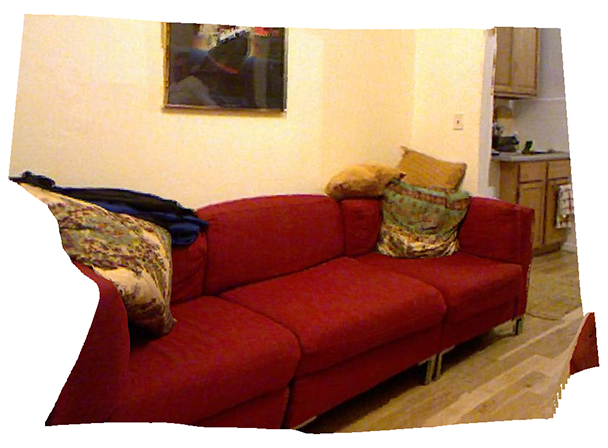}
    }
    \hfill
    \subfloat{
        \includegraphics[width=.23\textwidth]{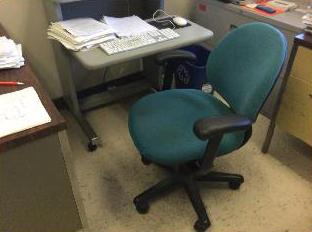}
    }\hspace{-5pt}
    \subfloat{
        \includegraphics[width=.23\textwidth]{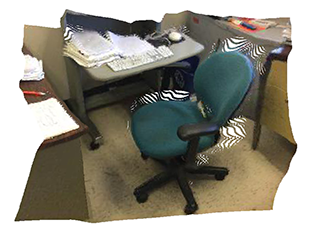}
    }
    \\ 
 \vspace{-5pt}
    \subfloat{
        \includegraphics[width=.23\textwidth]{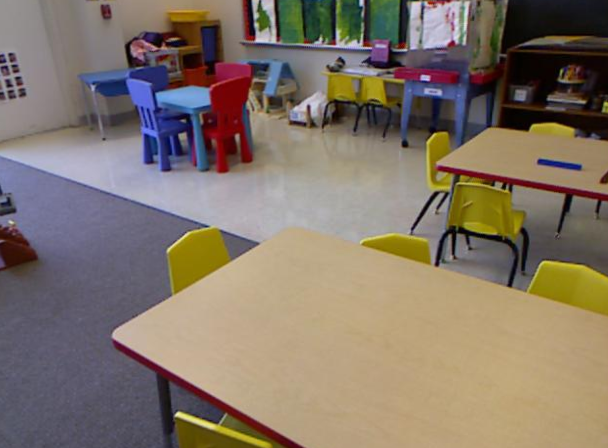}
    }\hspace{-5pt}
    \subfloat{
        \includegraphics[width=.23\textwidth]{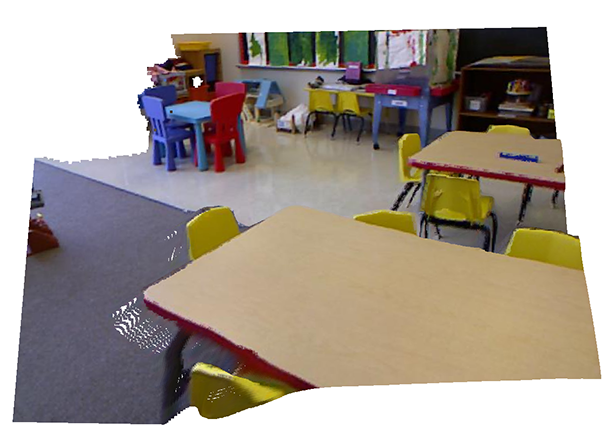}
    }
    \hfill
    \subfloat{
        \includegraphics[width=.23\textwidth]{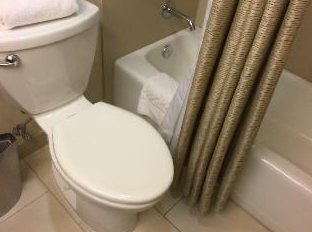}
    }\hspace{-5pt}
    \subfloat{
        \includegraphics[width=.23\textwidth]{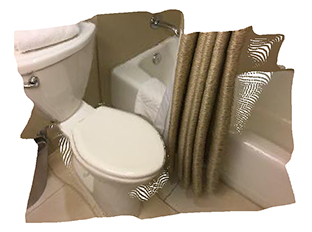}
    }
    \\ 
     \vspace{-5pt}
    \subfloat{
        \includegraphics[width=.23\textwidth]{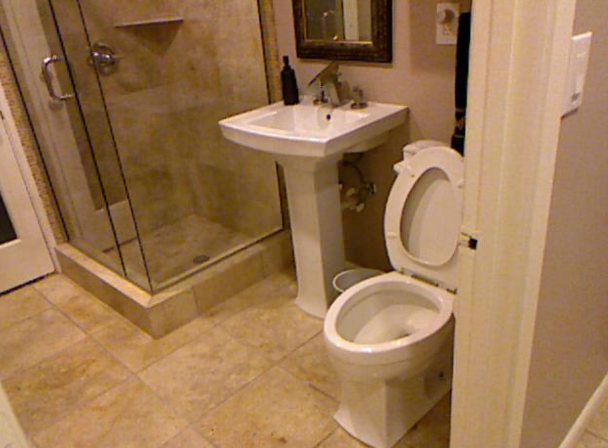}
    }\hspace{-5pt}
    \subfloat{
        \includegraphics[width=.23\textwidth]{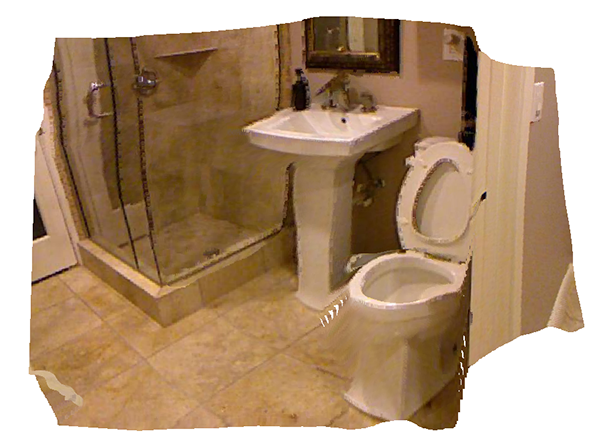}
    }
    \hfill
    \subfloat{
        \includegraphics[width=.23\textwidth]{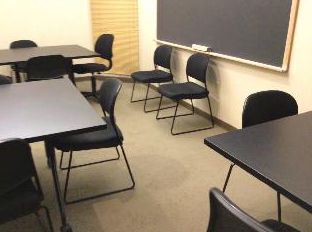}
    }\hspace{-5pt}
    \subfloat{
        \includegraphics[width=.23\textwidth]{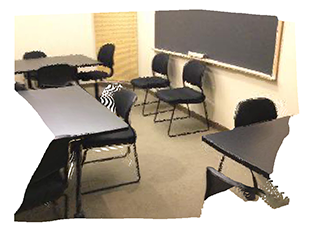}
    }
    \\ 
     \vspace{-5pt}
    \subfloat{
        \includegraphics[width=.23\textwidth]{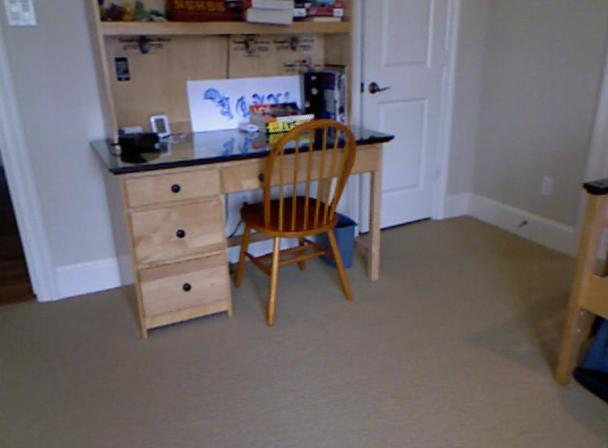}
    }\hspace{-5pt}
    \subfloat{
        \includegraphics[width=.23\textwidth]{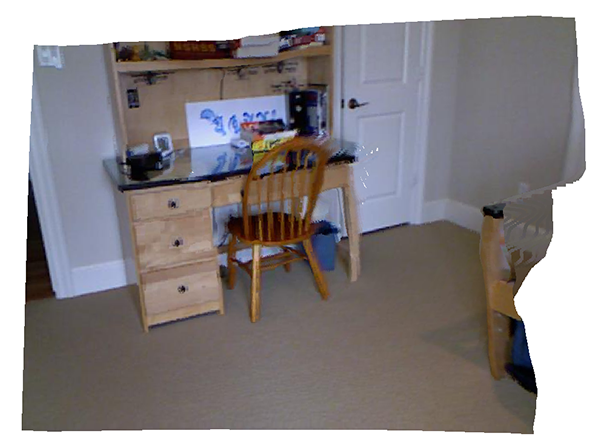}
    }
    \hfill
    \subfloat{
        \includegraphics[width=.23\textwidth]{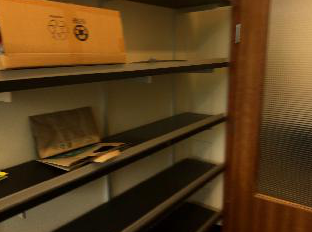}
    }\hspace{-5pt}
    \subfloat{
        \includegraphics[width=.23\textwidth]{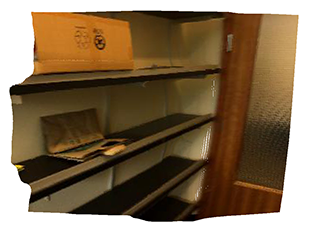}
    }
    \begin{flushleft}
        \caption{Visualisation of predicted point clouds. Left to right: input RGB image from NYUv2, 3D point cloud, input RGB image from ScanNet, 3D point cloud.\label{fig:3d}}
    \end{flushleft}
\end{figure}
Fig. \ref{fig:3d} shows the visualisation of point clouds generated by our model. We provide various samples on NYUv2 and ScanNet.

\vfill